\documentclass[11pt,a4paper]{article}

\usepackage{polyu-vclab}

\usepackage{lipsum}                % placeholder text, remove for real paper
\usepackage{amsmath}
\usepackage{booktabs}
\usepackage{multirow}
\usepackage[numbers,sort&compress]{natbib}
\usepackage{soul}
\usepackage{xcolor}
\soulregister\cite7
\soulregister\citep7
\soulregister\citet7
\soulregister\ref7
\soulregister\eqref7

\usepackage{hyperref}
\usepackage{url}
\usepackage{graphicx}
\usepackage[table]{xcolor}
\usepackage{array}
\usepackage{wrapfig}
\usepackage{longtable}
\usepackage{xltabular}
\newcolumntype{P}{>{\centering\arraybackslash}p{1.2cm}}
\definecolor{defrow}{RGB}{232,242,254}
\usepackage{booktabs}
\usepackage{adjustbox}
\setEyebrow{PolyU VCLab\,\textbullet\,Preprint 2026}

\setReportTag{Visual Computing Lab\,\textperiodcentered\,The Hong Kong Polytechnic University}

\papertitle{\centering OSWAM: An Effective World Action Model with One-Step Action Generation}

\paperauthors{\centering%
  Guowen Zhang\equalmark\affilmark{1}\quad
  Liyi Chen\equalmark\affilmark{1}\quad
  Xu Hu\affilmark{1}\quad
  Pengfei Wang\affilmark{1}\quad \\
  Jingfan Chen\affilmark{1}\quad
  Chenhang He\affilmark{1}\quad
  Lei Zhang\correspondmark\affilmark{1}%
}

\paperaffil{\centering%
  \affilmark{1}\,The Hong Kong Polytechnic University\quad
}

\papernotes{\centering%
  \equalmark\,Equal contribution.\quad
  \correspondmark\,Corresponding author. \\
  (\href{mailto:guowen.zhang@connect.polyu.hk}{guowen.zhang@connect.polyu.hk},
    \href{mailto:cslzhang@comp.polyu.edu.hk}{cslzhang@comp.polyu.edu.hk}).
}

\paperbadges{%
  \vclabbadgesolid{Project Page}{https://gwenzhang.github.io/OSWAM}\;%
  \vclabbadgesolid{Code}{https://github.com/PolyU-VCLab/OSWAM}\;%
}

\begin{document}
%==========================================================================

\maketitleVCLab

%-----------------------------------------------------------
% Abstract -- bordered box with red left bar (VCLab mission style)
%-----------------------------------------------------------
\begin{vclabAbstract}
\noindent\textbf{Abstract.}\;
Direct-policy World Action Models (WAM) have demonstrated the benefits of decoupling action generation from explicit future visual prediction for efficient inference in autonomous driving and embodied control.
However, they still require tens of denoising steps to achieve competitive action quality, incurring substantial latency and hindering real-time deployment.
Simply reducing the number of denoising steps will collapse the nonlinear multi-step denoising trajectory into coarse updates, substantially degrading action quality.
To reduce the denoising steps while enabling real-time action generation, we present an effective world action model with one-step action generation, namely \textbf{OSWAM}, for autonomous driving and embodied control.
We argue that it is the absence of structured action priors in random noise initialization that forces direct-policy WAMs to recover coherent action sequences through iterative denoising.
Hence, we construct imagination tokens to initialize the diffusion process, providing physical priors and action-conditioned representations.
We further introduce privileged information distillation with future conditioning to enhance future-aware action generation.
As a result, OSWAM can efficiently and effectively generate high quality actions in just one diffusion step.
Our experiments demonstrate that OSWAM achieves comparable or even better results in both autonomous driving and embodied control than direct-policy WAM and Joint-WAM that require tens of steps.
% Under compiled inference, OSWAM even achieves a real-time latency of $40$ \textit{ms} on an RTX 5090.
It obtains approximately $3\times$ speedup over direct-policy WAM baselines and supports real-time compiled inference with latency as low as $40$ms on an NVIDIA RTX 5090 GPU.
\end{vclabAbstract}


%==========================================================================
\section{Introduction}
%==========================================================================
Physical intelligence~\citep{agarwal2025cosmos,agarwal2026cosmos3}, including autonomous driving~\citep{survey_ad} and embodied control~\citep{fastwam}, holds the promise of freeing human workers from physical tasks that are dangerous, laborious, or tedious. 
A general-purpose physical agent should not only act from visual observations, but also anticipate the physical consequences of their actions. 
Existing Vision-Language-Action (VLA) models, including those standard models~\citep{black2024pi_0,intelligence2025pi_05,li2026drivevla-w0} and specialized end-to-end ones~\citep{liao2025diffusiondrive,chi2025diffusionpolicy_robot,li2025LaW}, typically predict actions directly without explicitly modeling the future evolution of the environment. 
This has motivated a shift toward World Action Models (WAMs)~\citep{fastwam,zhao2026simwam,li2026lingbot-va,liu2026driveva} for physical AI, which build on pretrained video generative models \citep{wan22,lingbot-world} to jointly predict future visual states and their corresponding actions.

\begin{figure}[t]
\centering
\includegraphics[width=1.0\textwidth]{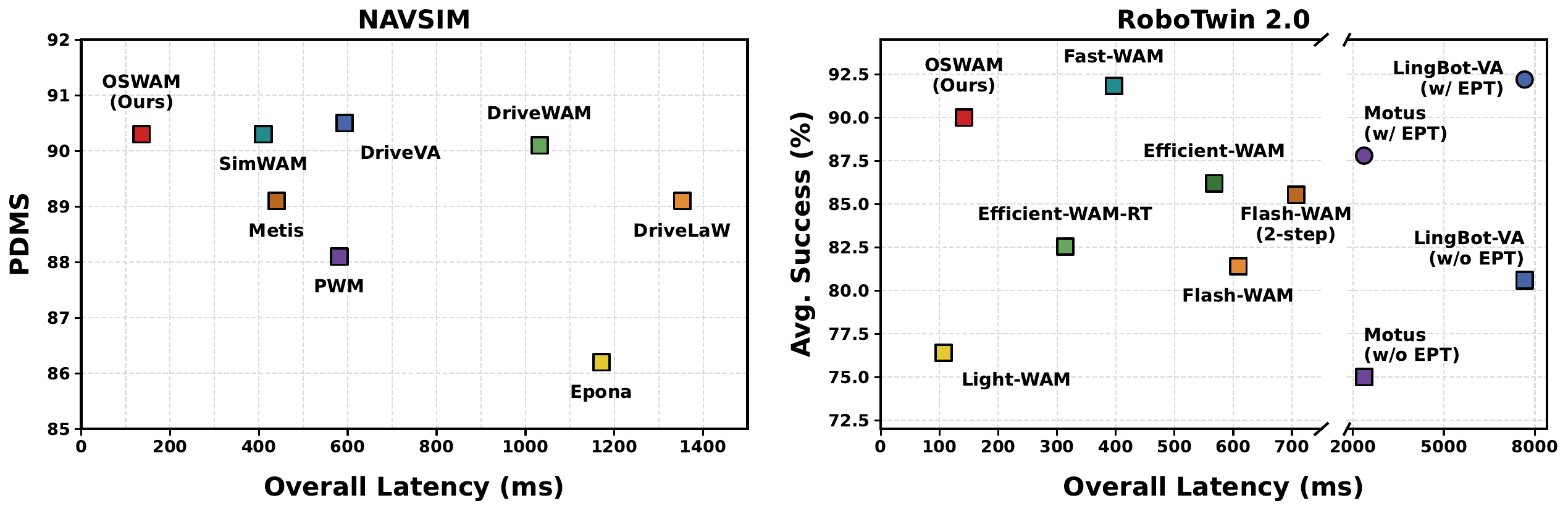}
\vspace{-7mm}
\caption{NAVSIM and Robotwin 2.0 inference efficiency-performance comparison. EPT denotes embodied pretraining. All latencies are measured on NVIDIA H20 GPU.}
\label{Fig::inference_performance}
\vspace{-4mm}
\end{figure}

% However, most existing WAMs, such as Joint-WAM \citep{bi2026motus,li2026lingbot-va}, incur prohibitive inference latency, rendering them impractical for latency-critical applications such as autonomous driving. 
% The imagine-then-execute paradigm~\citep{li2026lingbot-va}, which first generates future observations and then predicts actions conditioned on the imagined future, leads to substantial test-time latency.
However, WAMs follow the imagine-then-execute paradigm~\citep{bi2026motus,li2026lingbot-va} and incur prohibitive inference latency, rendering them impractical for latency-critical applications such as autonomous driving. 
Borrowing few-step techniques from image and video generation, some methods apply step distillation~\citep{akbari2026flashwam} to reduce this latency, but often at the expense of a substantial drop in control accuracy, as illustrated in Fig.~\ref{Fig::inference_performance}.
% \st{One possible strategy is to switch from the imagine-then-execute paradigm to direct-policy WAMs~\citep{fastwam,zhao2026simwam} to address this issue. }
{To address this issue, efforts~\citep{fastwam,zhao2026simwam} propose to  switch from the imagine-then-execute paradigm to direct-policy WAMs.  }
These methods retain video prediction as a training-time objective that shapes physically grounded representations, without synthesizing future observations at test time.
% This direct-policy formulation is more than $3\times$ faster than Joint-WAM or its inverse-dynamics-model (IDM) \citep{li2026lingbot-va} counterparts.
This direct-policy formulation is more than $3\times$ faster than the imagine-then-execute paradigm.
\begin{figure}[t]
\centering
\includegraphics[width=1.0\textwidth]{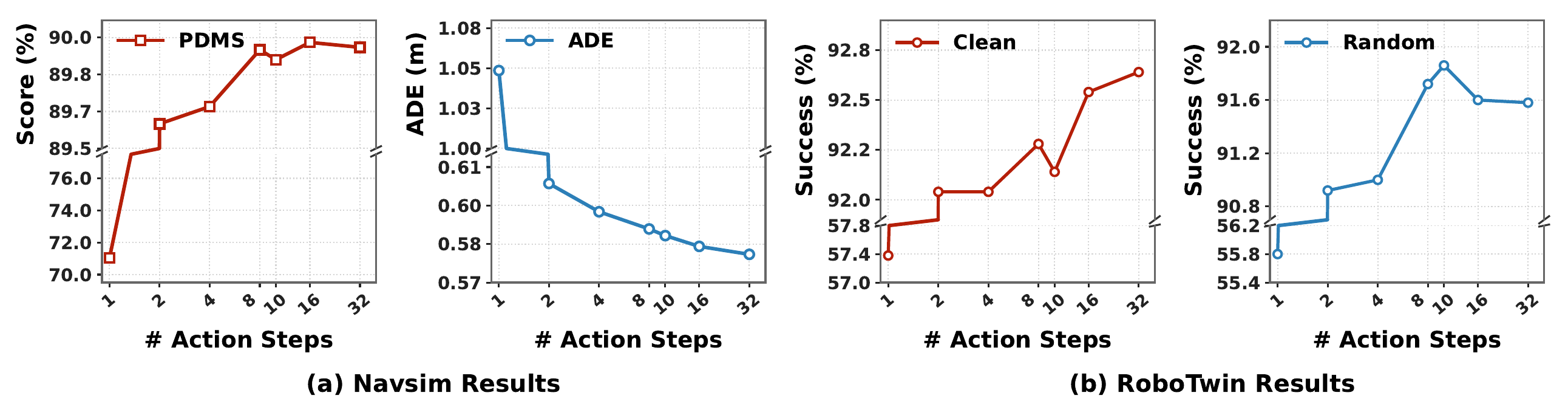}
\vspace{-6mm}
\caption{Quantitative comparison of the Fast-WAM ~\citep{fastwam} test-time scaling with different action denoising steps on (a) the autonomous driving dataset Navsim, and (b) embodied control dataset RoboTwin.}
\label{Fig::Insight}
\vspace{-4mm}
\end{figure}

% Despite this speedup, inference efficiency remains a key challenge for direct-policy WAMs~\cite{fastwam,zhao2026simwam}.
Despite this speedup, efficiently generating high-quality actions remains a key challenge for direct-policy WAMs.
% \st{require tens of denoising steps to achieve competitive action quality, whose latency remains too high for real-time applications such as autonomous driving.}
% \textcolor{red}{face two key limitations: inference inefficiency and limited effectiveness in future-aware control.
They require tens of denoising steps to achieve competitive action quality, resulting in latency that remains too high for real-time applications such as autonomous driving.
This bottleneck cannot be resolved by reducing the number of denoising steps.
Specifically, as shown in Fig.~\ref{Fig::Insight}, the performance degrades as the step count decreases, dropping by $20\%$ on NAVSIM (autonomous driving) and $40\%$ on RoboTwin (embodied control) in the one-step setting.
%These drops are substantially larger than the variation among different few-step budgets.
We attribute this to the lack of action structure in random noise: multiple denoising steps can gradually impose temporal and physical consistency, whereas one-step generation must recover a coherent action sequence in a single update.
Existing few-step accelerations, such as step distillation~\citep{akbari2026flashwam} and trajectory rectification~\citep{liu2022rectifiedflow}, still start from Gaussian noise and, therefore, do not supply the missing action prior.
%A more effective route is therefore to inject the missing structure at initialization.
% Action generation is low-dimensional and typically uses far fewer denoising steps than image or video synthesis, so introducing structured action priors is more appropriate than transplanting distillation designed for high-dimensional visual generation.
This motivates introducing structured action priors at initialization, providing an informed starting point for high-quality action generation in a single denoising step.

% \begin{wrapfigure}{r}{0.49\textwidth}
%     \centering
%     \includegraphics[width=\linewidth]{img/simwam_scores_ade_fde.pdf}
%     \vspace{-7mm}
%     \caption{Test-time scaling of Fast-WAM~\citep{fastwam}
%     with varying action denoising steps on (a) NAVSIM and (b) RoboTwin.}
%     \label{Fig::Insight}
%     \vspace{-3mm}
% \end{wrapfigure}
% \st{In addition, existing methods such as Direct-policy WAMs}
% \textcolor{red}{Second, direct-policy WAMs}
% \citep{fastwam,zhao2026simwam}
In addition, direct-policy WAMs~\cite{fastwam,zhao2026simwam} remain limited in their ability to exploit future dynamics for effective control.
They use a consistent first-frame-only interface during training and inference, relying on video co-training to encode future dynamics into the current representation.
However, compressing all future information into a single first-frame representation may provide insufficient temporal context for effective future-aware control.
% \st{Subsequent methods therefore reintroduce future features at inference, while reducing rollout cost through caching, sparse interaction}
{To address this limitation, subsequent methods reintroduce future features at inference and mitigate the associated rollout cost through caching, sparse interaction}
~\citep{zhao2026fasterwam_huake}, or compact latent representations~\citep{li2026efficientwam}.
% \st{Although these designs reduce the cost of future imagination, they still introduce future-dependent features at test time, increasing inference overhead.}
{Despite these efforts to reduce rollout cost, generating and incorporating future features at test time still incurs additional inference overhead.}
% and partially undermining the original goal of eliminating test-time future generation.
 
To address the aforementioned issues, we propose \textbf{OSWAM}, a world action model with one-step action generation for more effective physical control.
Our key observation is that previous one-step action generation methods fail mainly because the action diffusion processing is initialized from unstructured Gaussian noise.
To supply the missing action structure, we learn a compact set of \emph{imagination tokens} from the video Diffusion Transformer (DiT) and use them as structured start points for action denoising.
These tokens are far fewer than video first-frame tokens and are obtained in a single forward pass, so they add negligible computation, yet they interact with visual features to accumulate physically grounded, action-relevant context.
We then noise the imagination tokens to an intermediate timestep and decode them with the action DiT in one step, replacing the long denoising trajectory required by random initialization.
To enhance future-aware control, we train these imagination tokens with privileged-information distillation.
A teacher attends to visual tokens over the full future horizon, while the student imagination tokens, at test time, attend only to the first frame.
By distilling the teacher's privileged representations into the imagination tokens and subsequently into the action DiT, OSWAM transfers long-horizon context without reintroducing future-dependent features at inference.
This design preserves the first-frame-only interface of direct-policy WAMs, while enabling high-quality action generation in a single denoising step.

Our contributions are summarized as follows:
\begin{itemize}
\vspace{-2mm}
\item We propose OSWAM, a world action model that enables high-quality one-step action generation. 
It learns compact imagination tokens from the video DiT as structured start points for the action DiT. 
These tokens are obtained in a single forward pass with negligible overhead and are decoded in one denoising step.

\item We introduce privileged-information distillation to transfer long-horizon context into the imagination tokens. 
A teacher attends to visual tokens over the full future horizon, while the student remains first-frame-only at test time, preserving the interface of direct-policy WAMs without future-dependent features at inference.

\item Experiments on autonomous driving and embodied control show that OSWAM achieves competitive results with multi-step WAMs while being substantially faster.

\end{itemize}

%==========================================================================
\section{Related Work}
%==========================================================================

\subsection{World Action Models for Autonomous Driving}
Autonomous driving has recently shifted from modular pipelines with separate perception~\citep{zhang2024voxelmamba,zhang2024gga,he2024scatterformer,zhang2026bevdilation} and planning~\cite{cheng2024pluto} modules toward end-to-end frameworks~\citep{chitta2022transfuser}.
To further integrate future dynamics modeling, world models~\cite{wan22,chen2026omni3dedit,chen2026fast,wang2026one2scene} for autonomous driving have recently been extended into world action models (WAMs).
Existing methods fall into two categories: bird's-eye view (BEV)-based and video-generation-based world models. 
BEV-based approaches~\citep{li2025LaW,zheng2024occworld,li2025wote} forecast occupancy evolution and ego motion, predict future latent scene features, or infer future BEV layouts for online trajectory evaluation. 
They are computationally efficient but do not exploit large-scale video generative priors.
A complementary line of research leverages pretrained video generators and predicts the future in the image domain. 
These methods~\citep{shi2026drivewam,liu2026driveva,xia2026drivelaw} transfer such priors to driving, but jointly rolling out video features at test time remains inefficient. 
SimWAM~\citep{zhao2026simwam} follows the direct-policy design of Fast-WAM~\citep{fastwam} to avoid test-time video rollout, yet the remaining multi-step denoising is still impractical for onboard driving.
In contrast, our proposed OSWAM inherits large-scale video generative priors without test-time video rollout, and reduces action generation to a single denoising step.

% Unlike autonomous driving, where compact BEV world models remain a major alternative because planning is naturally grounded in a bird's-eye layout of roads and nearby agents, embodied control must model close-range object interactions and appearance changes. 
% Embodied WAMs are therefore built almost exclusively on pretrained video generators in the image domain. 
% Existing methods fall into two categories: joint imagine-then-execute WAMs and direct-policy WAMs. 
% Joint approaches~\citep{li2026lingbot-va,bi2026motus} transfer these priors by synthesizing future observations and then decoding actions, but jointly rolling out video features at test time remains inefficient. 
% Fast-WAM~\citep{fastwam} avoids test-time video rollout by predicting actions from current observations only, yet the remaining multi-step denoising still makes the latency impractical for closed-loop control. 
% In contrast, OSWAM inherits large-scale video generative priors without test-time video rollout, and reduces action generation to a single denoising step.

\subsection{World Action Models for Embodied Control} 
Unlike autonomous driving, where compact BEV representations naturally encode road topology and traffic interactions, embodied control needs to model fine-grained object interactions and appearance changes in close-range scenes. 
As these phenomena are naturally captured in video, embodied WAMs are predominantly built on pretrained video generators for image-domain future modeling in two ways: explicit future imagination and direct-policy inference.
{Explicit-imagination methods~\citep{li2026lingbot-va,bi2026motus} generate future visual representations at test time to guide action prediction, incurring substantial rollout latency. 
Direct-policy WAMs~\citep{fastwam} retain video prediction as a training signal but generate actions directly from current observations, avoiding test-time visual rollout; however, its iterative action denoising remains a bottleneck for high-frequency closed-loop control.}
In contrast, OSWAM preserves large-scale video-generative priors without test-time visual rollout and reduces action generation to a single denoising step.

\subsection{Effective World Action Models}
The main deployment bottleneck for WAMs lies in the high computational cost of diffusion-based video and action generators, whose  test-time latency is prohibitive for onboard driving and closed-loop manipulation.
%In embodied control, recent work has explored several ways to reduce this overhead.
Drawing one-step distillation from video generation, Flash-WAM~\citep{akbari2026flashwam} compresses joint video-action diffusion to a single denoising step per modality, but still synthesizes future video at inference.
Fast-WAM~\citep{fastwam} removes test-time video rollout by predicting actions from current observations only, yet retains iterative action denoising.
Light-WAM~\citep{li2026lightwam} further replaces the generative action expert with a lightweight fusion head for single-pass action prediction, but underperforms on challenging benchmarks such as RoboTwin.
%This direct-policy design has also been extended to autonomous driving: 
SimWAM~\citep{zhao2026simwam} follows Fast-WAM to drop test-time video rollout, but its multi-step action denoising keeps inference latency well above the budget required for onboard deployment.
Building on this video-rollout-free direct-policy formulation, OSWAM addresses the remaining action-generation bottleneck through one-step action denoising.

%==========================================================================
\section{Method}
%==========================================================================

\begin{figure*}[t!]
     \centering
    \includegraphics[width=0.95\textwidth]{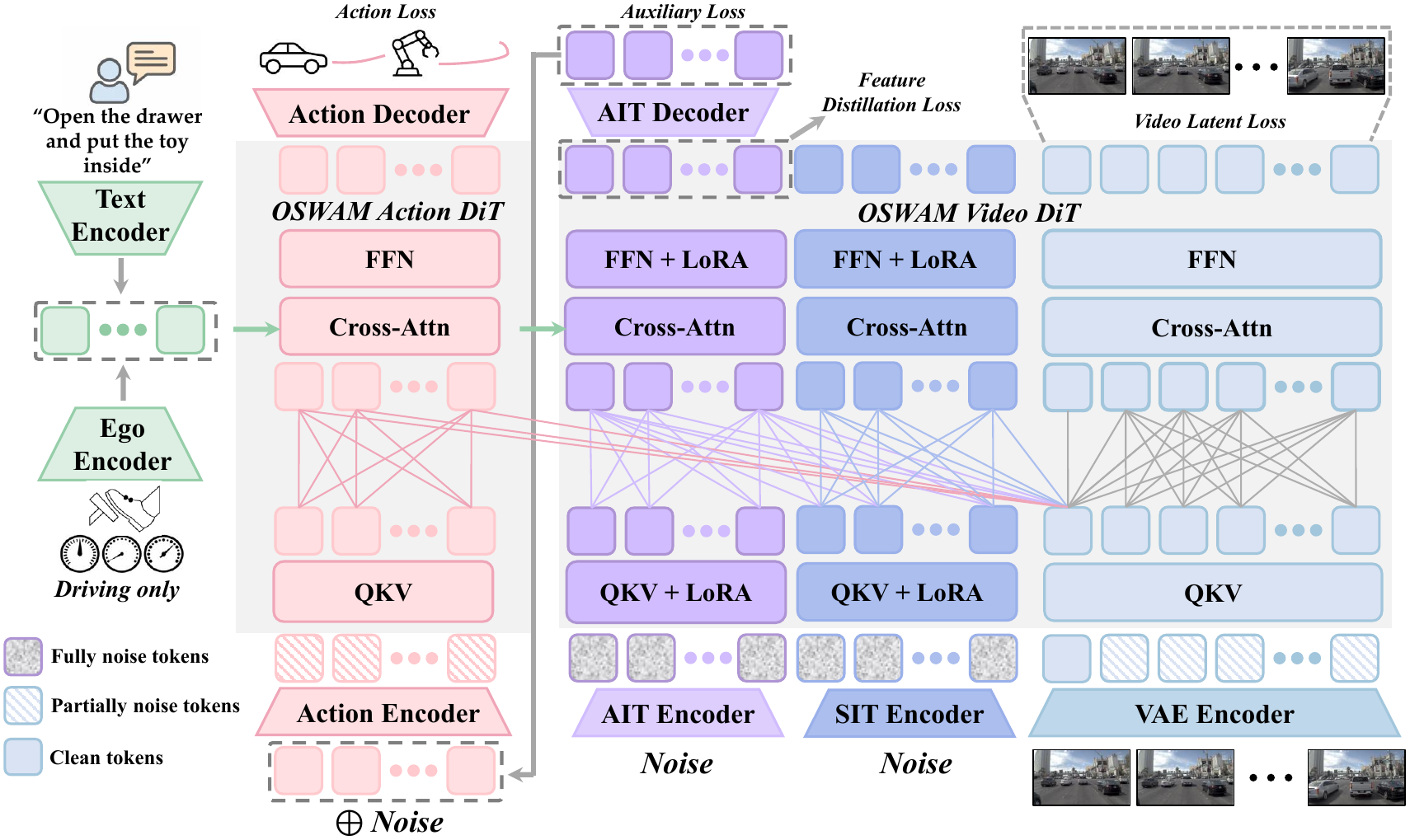}
    \vspace{-4mm}
     \caption{The overall architecture of OSWAM. A single Video DiT pass produces an observation-conditioned action prior with AIT and complementary scene context with SIT, providing a structured initialization for one-step Action DiT denoising. Privileged-information distillation injects future-informed representations during training, while inference uses only the current observation.
     }
    \label{fig:main}
    % \vspace{-6mm}
\end{figure*}

We present OSWAM, a direct-policy world action model that enables high-quality one-step action generation for autonomous driving and embodied control. 
The preliminary concepts associated with our method are provided in {Sec.~\ref{appedix:preliminary}}. In the following sections, we first present the overall architecture of OSWAM, then describe in detail the fundamental components of OSWAM, including imagination tokens, Role-aware 3D RoPE, and privileged-information distillation.

\subsection{Preliminary}
\label{appedix:preliminary}

A world action model predicts an action chunk by introducing future visual observations as an intermediate variable.
Let $o$ be the current observation, $l$ an optional command, $a_{1:H}$ an action chunk, and $v_{1:T}$ the future visual states.
Inverse-dynamics WAMs~\citep{li2026lingbot-va} first model the distribution of future visual states conditioned on the current observation and command, and then infer actions from the generated future:
\begin{equation}
p(a_{1:H}\mid o,l)
=\int p(v_{1:T}\mid o,l)\,p(a_{1:H}\mid o,l,v_{1:T})\,dv_{1:T}.
\end{equation}
Joint-WAMs~\citep{bi2026motus}, in contrast, model future video and actions within a shared generative process, $p(a_{1:H},v_{1:T}\mid o,l)$.
Both paradigms tightly couple control prediction with iterative future-video generation, which introduces substantial computational overhead.

Direct-policy WAMs~\citep{fastwam,zhao2026simwam} remove explicit test-time video generation by retaining video prediction only as a training objective. At inference, the video DiT produces a latent representation $z(o)$ in a single forward pass, and actions are predicted directly as:
\begin{equation}
p(a_{1:H}\mid o,l)=p(a_{1:H}\mid z(o)).
\end{equation}
However, action generation is iterative due to flow matching. Given $a_t=(1-t)a+t\epsilon$ with velocity target $\epsilon-a$, inference integrates the learned velocity field from noise ($t_K=1$) to data ($t_0=0$) as:
\begin{equation}
a_{t_{k-1}}
= a_{t_k} + (t_{k-1} - t_k)\, v_\theta(a_{t_k}, t_k).
\qquad k = K, \ldots, 1,
\end{equation}
High-quality actions typically require multiple velocity evaluations, making $K$ a remaining inference bottleneck.

% OSWAM aims to preserve this direct-policy interface while reducing action generation to a single velocity evaluation, $K=1$:
% \begin{equation}
% \hat{a} = a_{t_1} - v_\theta(a_{t_1}, 1).
% \end{equation}
% This one-step update replaces iterative action sampling, making direct-policy WAMs practical for latency-critical control, such as autonomous driving.
% OSWAM aims to preserve this direct-policy interface while reducing action generation to a single velocity evaluation, $K=1$:
% \begin{equation}
%     \hat{\mathbf{a}}
%     = \mathbf{a}_{\sigma_r}
%     - \sigma_r v_{\theta}(\mathbf{a}_{\sigma_r},\sigma_r),
% \end{equation}
% where $\sigma_r \in (0,1]$ is the normalized starting timestep.
% OSWAM uses a structured initialization at $\sigma_r<1$ to enable one-step action refinement, which makes direct-policy WAMs practical for latency-critical control, such as autonomous driving.

Our proposed OSWAM aims to preserve this direct-policy interface while reducing action generation to a single velocity evaluation, $K=1$.
A one-step update from an arbitrary time $t\in(0,1]$ is:
\begin{equation}
\hat{a} = a_t - t\, v_\theta(a_t, t).
\end{equation}
When the start is pure noise, $t=1$ and $\hat{a}=a_1-v_\theta(a_1,1)$.
OSWAM instead starts from a structured intermediate state with $t<1$, which makes this single update sufficient for applications that require latency-critical control, such as autonomous driving.

\subsection{Overall Architecture}
\label{sec:overall_architecture}
An overview of OSWAM is shown in Fig.~\ref{fig:main}. 
As in previous works~\citep{fastwam,zhao2026simwam}, we consider predicting an action chunk $\mathbf{a}_{1:H}$ from the current frame $f_0$ and an optional task instruction $l$, with ego-state information additionally provided for driving tasks. 
We follow Fast-WAM in the world model backbone configuration and direct-policy setting. 
Specifically, the Video DiT uses the Wan2.2-5B backbone~\citep{wan22}, accompanied by its pretrained T5 text encoder and video VAE.
The Action DiT is initialized by interpolating and downsampling the pretrained Video DiT weights.
For autonomous driving, we add an ego encoder to encode the ego state and the driving command.
The task languages and ego states are both provided to all tokens through cross-attention.
The full model is organized as a Mixture-of-Transformer (MoT) architecture with shared attention between the video and action blocks.
We also follow the direct-policy protocol: video modeling is used only as a training signal, and future video is not rolled out at inference.

Unlike prior direct-policy methods, which use the Video DiT at inference solely to encode the first-frame visual feature for Action DiT through shared MoT attention, OSWAM additionally introduces two groups of imagination tokens into the Video DiT: \emph{Action Imagination Tokens} (AIT) and \emph{Scene Imagination Tokens} (SIT).
As shown in Fig.~\ref{fig:main}, both AIT and SIT are initialized from noise in the Video DiT and interact with the first-frame visual features. 
The resulting AIT representation then replaces the unstructured Gaussian initialization of the Action DiT, providing a compact, structured start point for action denoising.
% SIT serve as additional latent slots for aggregating scene information and learning from privileged supervision.
SIT provides additional latent slots that aggregate scene context for AIT; they are updated indirectly through privileged distillation and the auxiliary action loss on AIT.
Together, AIT and SIT comprise far fewer tokens than the first-frame visual input and therefore introduce only a small computational overhead during inference.
To strengthen future-aware control under this first-frame-only interface, we introduce privileged-information distillation.
% A privileged teacher with the same architecture accesses both the current and future frames, learning future-informed representations from the scene evolution.
The privileged teacher comprises a Video DiT with AIT that attends to visual tokens over the full clip, while the AIT in OSWAM attends only first-frame visual context.
By aligning OSWAM's intermediate imagination token features with those of the teacher, we transfer this privileged information to improve action generation, while OSWAM remains first-frame-only at inference.

\begin{figure*}[t!]
     \centering
    \includegraphics[width=1\textwidth]{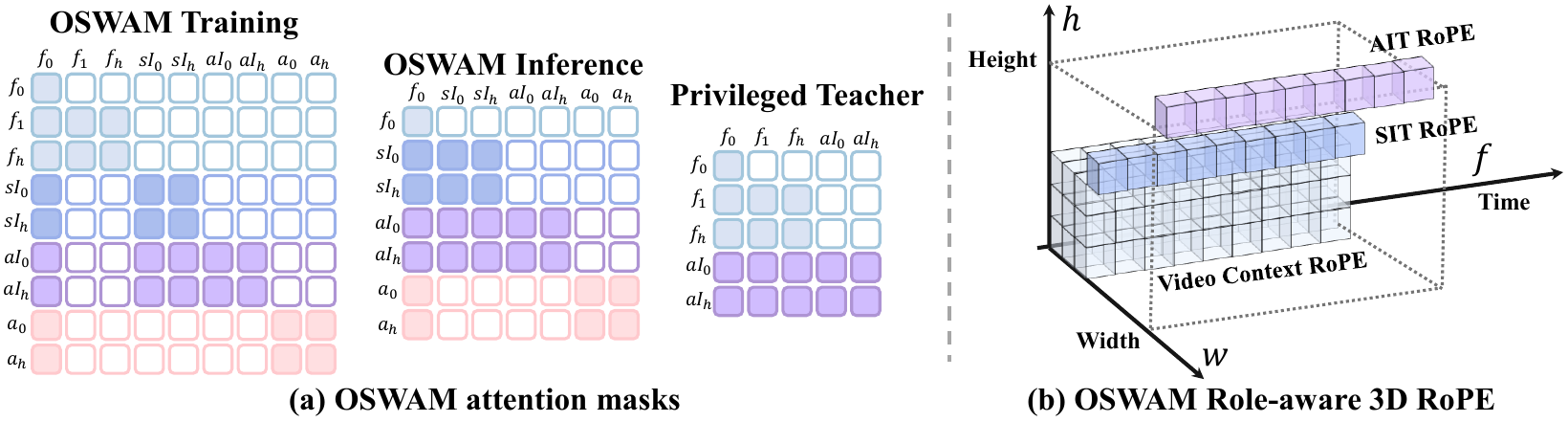}
     \vspace{-5mm}
     \caption{(a) The attention mask of OSWAM and the privileged teacher. (b) Role-aware RoPE explicitly disentangles token groups using unique positional encodings. $sI$ is SIT and $aI$ is AIT.
     }
    \label{fig:attn mask}
    % \vspace{-6mm}
\end{figure*}

\subsection{Imagination Tokens}
\label{sec:imagination_tokens}

% Direct-policy WAMs initialize action generation from Gaussian noise and iteratively refine action tokens through MoT attention with visual features.
% This avoids explicit future-video generation, but an unstructured initialization typically still requires multiple denoising steps.
% To enable one-step action generation, we introduce two groups of compact imagination tokens.
Direct-policy WAMs eliminate future video generation at inference but still require multi-step denoising. To enable one-step action generation, we introduce two groups of imagination tokens.

\textbf{Action and Scene Imagination Tokens.}
As mentioned in the last subsection, we augment the Video DiT input with two groups of noise-initialized latent tokens: AIT and SIT.
% with token counts $N_A$ and $N_S$, respectively.
% As shown in the Figure~\ref{fig:attn mask} (a), both groups interact with first-frame visual tokens under a structured attention mask and receive task conditioning through cross-attention.
We adapt the Video DiT using LoRA~\citep{hu2021lora} to support these additional tokens.
Their combined token count is much smaller than that of the first-frame visual input tokens, introducing only a small computational overhead during inference.

During training, video tokens follow the standard diffusion process with randomly sampled noise levels, whereas AIT are constructed by encoding an initial action noise sample $\boldsymbol{\epsilon}\sim\mathcal{N}(0,I)$ and assigned the fixed timestep $t_{\max}$.
Within the Video DiT, AIT attends to first-frame visual tokens, SIT, and other AIT, while receiving task conditioning through cross-attention.
These interactions integrate visual evidence and scene context into AIT.
An auxiliary decoder maps the resulting AIT representations to an action velocity $\hat{\mathbf{v}}^{A}$, yielding a structured action prior $\bar{\mathbf{a}}=\boldsymbol{\epsilon}-\hat{\mathbf{v}}^{A}$.
% An auxiliary decoder maps the resulting AIT representations to a coarse action estimate $\bar{\mathbf{a}}$ in a single update.
We then detach this prior and renoise it with the same initial noise that was encoded into AIT:
\begin{equation}
    \mathbf{a}^{\mathrm{init}}_{\tau_r}
    = (1-\sigma_r)\operatorname{sg}(\bar{\mathbf{a}})
    + \sigma_r\boldsymbol{\epsilon},
    \qquad \sigma_r = \tau_r/t_{\max},
\end{equation}
where $\sigma_r$ controls the noise level and $\operatorname{sg}$ prevents gradients from propagating through the structured action prior.
Conditioned on first-frame visual features and task inputs, the Action DiT predicts a refinement velocity $\hat{\mathbf{v}}^{\mathrm{act}}$ and performs the single denoising step:
\begin{equation}
    \hat{\mathbf{a}}
    = \mathbf{a}^{\mathrm{init}}_{\tau_r}
    - \sigma_r\hat{\mathbf{v}}^{\mathrm{act}}.
\end{equation}
% Its training target is $(\mathbf{a}^{\mathrm{init}}_{\tau_r}-\mathbf{a}_{\mathrm{gt}})/\sigma_r$, allowing the refinement step to remove noise and correct errors in the coarse action estimate.
Its training target is $(\mathbf{a}^{\mathrm{init}}_{\tau_r}-\mathbf{a}_{\mathrm{gt}})/\sigma_r$, enabling the Action DiT to refine the structured initialization into the target action in a single denoising step.

% SIT provide complementary latent slots for aggregating scene information and receiving supervision through privileged-information distillation.
% Their intended role resembles that of registers that collect contextual information, with their contents computed dynamically from noise and visual inputs.
% The teacher constructs these representations with access to future observations, while the student learns to infer them from the first frame alone.
% The roles of AIT and SIT are differentiated through their RoPE assignments and supervision: AIT receive an auxiliary action prediction loss, whereas SIT are supervised by feature distillation from the privileged teacher.
% We describe their positional assignments next.
To complement the action prior encoded by AIT, SIT acts as latent context slots that aggregate scene context for AIT to attend to.
They are initialized from Gaussian noise independent of the AIT, projected into the Video DiT input space through a separate encoder, and assigned the timestep $t_{\max}$.
As shown in Fig.~\ref{fig:attn mask} (a), under the structured attention mask, SIT attends to first-frame visual tokens and other SIT, while AIT additionally attends to SIT to access the aggregated scene context.
% This allows SIT to serve as contextual registers that supply scene information for action prediction.
SIT is learned indirectly through the auxiliary loss and feature distillation on AIT, with gradients flowing through the attention connections from AIT to SIT.

\textbf{Role-aware 3D RoPE.}
As shown in Fig.~\ref{fig:attn mask} (b), we introduce Role-aware 3D RoPE to reflect the distinct functions of video tokens, AIT, and SIT within a shared positional coordinate system.
Video tokens represent local visual content and retain their original spatiotemporal coordinates.
AIT encodes action structure and is assigned spatial coordinates outside the video grid to distinguish them from visual tokens.
SIT serves as scene context aggregators and is anchored at the grid center, providing a common spatial reference for gathering information across the scene:
% Let $H_v$ and $W_v$ denote the spatial dimensions of the video token grid. We assign coordinates as
% \begin{align}
%     \mathbf{p}^{V}_{t,h,w} &= (t,h,w), \\
%     \mathbf{p}^{A}_{n} &= (t_n,H_v,W_v), \\
%     \mathbf{p}^{S}_{n}
%     &= \left(t_n,\frac{H_v-1}{2},\frac{W_v-1}{2}\right).
% \end{align}
\begin{equation}
\mathbf{p}^{V}_{t,h,w}=(t,h,w),\quad
\mathbf{p}^{A}_{n}=(t_n,H_v,W_v),\quad
\mathbf{p}^{S}_{n}=\Bigl(t_n,\tfrac{H_v-1}{2},\tfrac{W_v-1}{2}\Bigr).
\end{equation}
where $H_v$ and $W_v$ are the video token grid dimensions.
For each action position $n$, the corresponding AIT and SIT share the temporal coordinate $t_n=n s_a/(s_v d_t)$.
Here, $s_a$ and $s_v$ are the action and video sampling strides, and $d_t$ is the VAE temporal downsampling factor.
This converts action positions to the video latent timeline.
% The shared temporal coordinates align AIT and SIT in time, while their distinct spatial anchors distinguish their roles.
% The shared temporal coordinates differ in their spatial anchors, preserving temporal correspondence while providing positional cues for structured initialization prediction and scene aggregation, respectively.
Thus, AIT and SIT share temporal coordinates but use distinct spatial anchors, preserving temporal alignment while providing role-specific positional cues for action initialization and scene context aggregation, respectively.
% Here, $t$ denotes temporal position rather than diffusion timestep, and $t_n=n s_a/(s_v d_t)$ aligns imagination tokens with the video timeline, where $s_a$ and $s_v$ are the action and video sampling strides and $d_t$ is the VAE temporal downsampling factor.
% AIT and SIT share temporal coordinates but differ in their spatial anchors, preserving temporal correspondence while providing positional cues for structured initialization prediction and scene aggregation, respectively.

\subsection{Privileged-Information Distillation}
\label{sec:privileged_distillation}

Although imagination tokens provide a structured action prior, long-horizon context is only implicitly encoded in the first-frame representation. 
We therefore introduce privileged-information distillation to transfer representations learned from privileged future observations into OSWAM's AIT, strengthening their predictive capability without requiring future observations at inference.

We separately train a Video DiT teacher with auxiliary AIT, following the same training settings as OSWAM except for the token groups and attention masks.
As shown in Fig.~\ref{fig:attn mask} (a), the teacher contains video tokens and AIT, with AIT attending to the full video clip.
OSWAM additionally includes SIT, and its AIT attends to first-frame visual tokens, SIT, and other AIT, while SIT attends only to first-frame visual tokens and other SIT.
% The first-frame tokens cannot attend to future frames, preventing future information from reaching the student's imagination tokens indirectly.

% During distillation, we freeze the teacher and perform a forward pass with partially noised future video latents at a fixed timestep $\tau_v^{T}$, keeping the first-frame latents clean.
During distillation, we freeze the teacher and provide it with ground-truth video latents over the full future horizon.
The first-frame latents remain clean, while all future-frame latents are corrupted with noise at a fixed diffusion timestep $\tau_v^{T}$ before the teacher forward pass.
The teacher's AIT remains initialized from pure noise at $t_{\max}$ and acquires privileged information by attending to these future video tokens.
OSWAM retains its standard training procedure, using a randomly sampled video timestep and the fixed AIT timestep $t_{\max}$.
The teacher and OSWAM share the same video noise and action noise samples, respectively.
We then align OSWAM's AIT features with the corresponding teacher features at a set of Video DiT blocks $\mathcal{J}$, providing multi-depth distillation supervision. 
SIT is supervised indirectly through their attention connections to AIT. 
At inference, the teacher is discarded, preserving OSWAM's first-frame-only interface without additional computation.

\subsection{Training Objective}
\label{sec:training_objective}

OSWAM jointly learns visual dynamics, structured action DiT start points, and one-step action denoising.
The Video DiT is supervised by video flow matching, while its AIT representations receive complementary action and privileged feature supervision to learn the structured initialization consumed by the Action DiT:
\begin{equation}
    \mathcal{L}_{\mathrm{vid}}
    = \mathbb{E}\bigl[
        \|\hat{\mathbf{v}}^{V}
        -(\boldsymbol{\epsilon}^{V}-\mathbf{z}^{+})\|_2^2
    \bigr],
\end{equation}
where $\mathbf{z}^{+}$ denotes clean future video latents and $\boldsymbol{\epsilon}^{V}$ is the video noise.
To learn an observation-conditioned action prior, the AIT decoder is supervised with the action velocity target at $t_{\max}$:
\begin{equation}
    \mathcal{L}_{\mathrm{AIT}}
    = \mathbb{E}\bigl[
        \|\hat{\mathbf{v}}^{A}
        -(\boldsymbol{\epsilon}-\mathbf{a}_{\mathrm{gt}})\|_2^2
    \bigr].
\end{equation}
Privileged feature distillation further guides the prior by aligning the OSWAM's AIT representations with teacher across the selected blocks $\mathcal{J}$:
% \begin{equation}
%     \mathcal{L}_{\mathrm{dist}}
%     = \mathbb{E}\!\left[
%         \frac{1}{|\mathcal{J}|}
%         \sum_{j\in\mathcal{J}}
%         \operatorname{MSE}\!\left(
%         \mathbf{h}^{\mathrm{stu}}_{j},
%         \operatorname{sg}(\mathbf{h}^{\mathrm{tea}}_{j})
%         \right)\right],
% \end{equation}
\begin{equation}
    \mathcal{L}_{\mathrm{dist}}
    = |\mathcal{J}|^{-1}\mathbb{E}\bigl[
        \sum\nolimits_{j\in\mathcal{J}}
        \|\mathbf{h}^{\mathrm{stu}}_{j}
        -\operatorname{sg}(\mathbf{h}^{\mathrm{tea}}_{j})\|_2^2
    \bigr],
\end{equation}
where $\mathbf{h}^{\mathrm{stu}}_{j}$ and $\mathbf{h}^{\mathrm{tea}}_{j}$ denote corresponding AIT features after block $j$. Together, $\mathcal{L}_{\mathrm{AIT}}$ and $\mathcal{L}_{\mathrm{dist}}$ support learning an observation-conditioned structured initialization for the Action DiT.
Given $\mathbf{a}^{\mathrm{init}}_{\tau_r}$ constructed from the AIT prediction, the Action DiT learns to predict the demonstrated action chunks in one denoising step:
\begin{equation}
    \mathcal{L}_{\mathrm{act}}
    = \mathbb{E}\bigl[
        \|\hat{\mathbf{v}}^{\mathrm{act}}
        -\sigma_r^{-1}
        (\mathbf{a}^{\mathrm{init}}_{\tau_r}
        -\mathbf{a}_{\mathrm{gt}})\|_2^2
    \bigr].
\end{equation}
Scaling the action residual by $\sigma_r^{-1}$ ensures that one Euler update with step size $\sigma_r$ directly moves $\mathbf{a}^{\mathrm{init}}_{\tau_r}$ to the ground-truth action $\mathbf{a}_{\mathrm{gt}}$.
The overall objective is:
\begin{equation}
    \mathcal{L}
    = \mathcal{L}_{\mathrm{vid}}
    + \mathcal{L}_{\mathrm{AIT}}
    + \lambda_{\mathrm{dist}}\mathcal{L}_{\mathrm{dist}}
    + \mathcal{L}_{\mathrm{act}}.
\end{equation}

%==========================================================================
\section{Experiments}
%==========================================================================

\begin{table}[t!]
  \caption{Comparison with state-of-the-art planners on the NAVSIM \texttt{navtest}. $N^{v}$ and $N^{a}$ denote the number of video and action \textbf{denoising steps}. $^{\dagger}$SimWAM is reported without reinforcement learning. $^{\ddagger}$DriveVA is reported without CARLA mix training. $^{*}$Latency reproduced by us.}
  \label{tab:navsim}
  \centering
  \setlength\tabcolsep{2.4mm}
  \renewcommand{\arraystretch}{1.0}
  \resizebox{0.9\linewidth}{!}{%
  \begin{tabular}{l c c c c c c c c c}
    \toprule
    Method & $N^{v}$ & $N^{a}$ & NC$\uparrow$ & DAC$\uparrow$ & EP$\uparrow$ & TTC$\uparrow$ & C$\uparrow$ & PDMS$\uparrow$ & Latency$\downarrow$ \\
    \midrule
    Human Agent & -- & -- & 100.0 & 100.0 & 87.5 & 100.0 & 99.9 & 94.8 & -- \\
    \midrule
    \multicolumn{10}{l}{\textit{Traditional E2E and VLA-based planners}} \\
    UniAD~\citep{hu2023uniad} & -- & -- & 97.8 & 91.9 & 78.8 & 92.9 & 100.0 & 83.4 & -- \\
    TransFuser~\citep{chitta2022transfuser} & -- & -- & 97.7 & 92.8 & 79.2 & 92.8 & 100.0 & 84.0 & -- \\
    % ARTEMIS~\citep{artemis2025} & -- & -- & 98.3 & 95.1 & 81.4 & 94.3 & 100.0 & 87.0 & -- \\
    % WorldRFT~\citep{yang2026worldrft} & -- & -- & 97.8 & 96.8 & 81.7 & 94.0 & 100.0 & 87.8 & -- \\
    DiffusionDrive~\citep{liao2025diffusiondrive} & -- & -- & 98.2 & 96.2 & 82.2 & 94.7 & 100.0 & 88.1 & -- \\
    WoTE~\citep{li2025wote} & -- & -- & 98.5 & 96.8 & 81.9 & 94.9 & 99.9 & 88.3 & -- \\
    % SeerDrive~\citep{zhang2026seerdrive} & -- & -- & 98.4 & 97.0 & 83.2 & 94.9 & 99.9 & 88.9 & -- \\
    % \midrule
    % \multicolumn{10}{l}{\textit{VLM-based planners}} \\
    UniWorldVLA~\citep{liu2026uniworldvla} & -- & -- & 98.7 & 96.7 & 83.2 & 96.1 & 100.0 & 89.4 & -- \\
    DriveDreamer-Policy~\citep{zhou2026drivedreamer} & -- & -- & 98.4 & 97.1 & 83.5 & 95.1 & 100.0 & 89.2 & -- \\
    % Vega~\citep{zuo2026vega} & -- & -- & 98.9 & 95.3 & 81.6 & 96.1 & 100.0 & 87.9 & -- \\
    % ImagiDrive~\citep{li2025imagidrive} & -- & -- & 98.6 & 96.2 & 80.5 & 94.5 & 100.0 & 87.4 & -- \\
    AutoVLA~\citep{zhou2026autovla} & -- & -- & 98.4 & 95.6 & 81.9 & {98.0} & 99.9 & 89.1 & -- \\
    % ReCogDrive~\citep{} & -- & -- & 97.9 & 97.3 & \textbf{87.3} & 94.9 & 100.0 & 90.8 & -- \\
    % ExploreVLA~\citep{sheng2026explorevla} & -- & -- & 98.8 & 98.4 & 83.5 & 96.5 & 99.9 & 90.4 & -- \\
    DriveVLA-W0~\citep{li2026drivevla-w0} & -- & -- & 98.7 & {99.1} & 83.3 & 95.3 & 99.3 & 90.2 & -- \\
    % SGDrive~\citep{li2026sgdrive} & -- & -- & 98.6 & 97.8 & 85.8 & 96.2 & 100.0 & 91.1 & -- \\
    \midrule
    \multicolumn{10}{l}{\textit{Imagine-then-execute-based WAM methods}} \\
    Epona~\citep{zhang2025epona} & -- & 100 & 97.9 & 95.1 & 80.4 & 93.8 & 99.9 & 86.2 & 1171 \\
    PWM~\citep{zhao2026pwm} & -- & -- & 98.6 & 95.9 & 81.8 & 95.4 & 100.0 & 88.1 & 581 \\
    DriveLaW~\citep{xia2026drivelaw} & 1 & 5 & {99.0} & 97.1 & 81.3 & 96.7 & 100.0 & 89.1 & 1353 \\
    DriveVA$^{\ddagger}$~\citep{liu2026driveva} & 2 & 2 & 99.0 & 97.3 & 83.2 & 98.4 & 100.0 & 90.5 & 593$^{*}$\\
    DriveWAM~\citep{shi2026drivewam} & 3 & 10 & 98.3 & 98.1 & 84.3 & 95.2 & 100.0 & 90.1 & 1032 \\
    \midrule
    \multicolumn{10}{l}{\textit{Direct policy-based WAM planners}} \\
    SimWAM$^{\dagger}$~\citep{zhao2026simwam} & -- & 10 & 98.7 & 98.0 & 83.9 & 95.9 & 100.0 & 90.3 & 410 \\
    Metis~\citep{li2026metis} & -- & 10 & 98.3 & 97.1 & 83.4 & 94.7 & 100.0 & 89.1 & 440$^{*}$ \\
    \midrule
    \textbf{OSWAM} ({Ours}) & -- & 1 & 98.6 & 98.0 & 84.3 & 95.6 & 100.0 & {90.3} & 136 \\
    \bottomrule
  \end{tabular}%
  }
  \vspace{-3mm}
\end{table}

\subsection{Datasets and Evaluation Metrics}
\textbf{Autonomous Driving.}
For autonomous driving, we evaluate OSWAM on the NAVSIM~\citep{dauner2024navsim}, which provides real-world driving data sampled at 2Hz, with eight cameras covering a 360-degree
field of view and LiDAR observations.
The dataset comprises 1192 scenarios for training and 136 scenarios for testing.
NAVSIM introduces practical and reliable metrics that effectively bridge open-loop and closed-loop evaluation.
Specifically, NAVSIM evaluates model performance using the Predictive Driver Model Score (PDMS), which is calculated based on five factors: No At-Fault Collision (NC), Drivable Area Compliance (DAC), Time-to-Collision (TTC), Comfort (Comf.), and Ego Progress (EP). We report PDMS, which is calculated as $\mathrm{PDMS}=\mathrm{NC}\times\mathrm{DAC}\times({5\times \mathrm{EP} + 5\times \mathrm{TTC} + 2\times \mathrm{C}})/{12}.$
% $PDMS=\mathrm{NC}\times \mathrm{DAC}\times\frac{5,\mathrm{EP}+5,\mathrm{TTC}+2,\mathrm{Comf.}}{12}$.
% $\mathrm{PDMS}=\mathrm{NC}\times\mathrm{DAC}\times({5\times \mathrm{EP} + 5\times \mathrm{TTC} + 2\times \mathrm{C}})/{12}.$

\textbf{Embodied Control.}
For embodied control, we evaluate OSWAM on RoboTwin 2.0~\citep{chen2025robotwin}, which is a challenging bimanual manipulation benchmark featuring more than 50 tasks that require coordinated dual-arm control. We adopt the multi-task setting of Fast-WAM, using data from over 50 tasks. The training data consist of 2,500 demonstrations from clean scenes and 25,000 demonstrations collected with heavy scene randomization. Performance is evaluated by the average success rate over 100 trials per task under both clean and randomized conditions.

\subsection{Implementation Details}
\label{sec:implementation_details}

Following Fast-WAM, we initialize OSWAM from Wan2.2-5B and adopt the Action DiT with a reduced hidden dimension of 1024.
The imagination-token LoRA adapters use rank 8 and $\alpha=4$.
For NAVSIM, we follow \citep{zhao2026simwam} to use a single front camera at $384\times672$ resolution and predict eight waypoints over four seconds at 2Hz.
We use the current frame and eight future frames in training.
% together with eight future frames during training.
For RoboTwin, we concatenate the head and two wrist views into a $384\times320$ image and use an action horizon of 32, with nine video frames sampled at a temporal stride of 4.
For temporal RoPE, we set $s_a=1$ and $d_t=4$ on both benchmarks, with $s_v=1$ for NAVSIM and $s_v=4$ for RoboTwin.
AIT and SIT therefore use $t_n=n/4$, ($n=1,\ldots,8$) on NAVSIM and $t_n=n/16$, ($n=1,\ldots,32$) on RoboTwin; video latents use temporal coordinates $\{0,1,2\}$ in training.

We train with AdamW using a learning rate of $10^{-4}$, weight decay of $0.01$, and cosine decay with 5\% linear warmup.
NAVSIM training runs for 40,000 steps with a global batch size of 256, while RoboTwin training runs for 29,335 steps with a global batch size of 1024.
For distillation, the frozen teacher uses a video timestep of 200, and feature alignment is applied after Video DiT blocks $[8, 16, 24, 30]$.
We set $\lambda_{\mathrm{dist}}=10$.
Distillation starts after 500 steps, with its weight increasing linearly over the next 500 steps.
Across both benchmarks, we set the AIT diffusion timestep to $t_{\max}=1000$ and the re-noising timestep for Action DiT initialization to $\tau_r=500$ ($\sigma_r=0.5$).
During inference, we use a single Action DiT denoising step. Unless otherwise specified, all latency measurements are conducted on a single NVIDIA H20 GPU and Intel Xeon Max 9468 CPU.

\begin{table}[t]
\centering
\caption{Success rates on RoboTwin 2.0 simulation (Clean and Randomized splits, 50 tasks). EPT denotes embodied pretraining.}
\resizebox{0.8\textwidth}{!}{%
\begin{tabular}{lccccccc}
\toprule
Method & $N^v$ & $N^a$ & EPT & Clean & Rand. & Average & Latency (ms) \\
\midrule
\multicolumn{8}{l}{\textit{VLA-based methods}}\\
$\pi_0$~\citep{black2024pi_0}                          & -- & -- & w/ & 65.92 & 58.40 & 62.20  & -- \\
$\pi_{0.5}$~\citep{intelligence2025pi_05}                     & -- & -- & w/ & 82.74 & 76.76 & 79.80  & -- \\
X-VLA~\citep{zheng2510xvla}                           & -- & -- & w/ & 72.90  & 72.80  & 72.80  & -- \\
\midrule
\multicolumn{8}{l}{\textit{Imagine-then-execute-based WAM methods}}\\
Motus~\citep{bi2026motus}                          & -- & -- & w/ & 88.66 & 87.02 & 87.80  & 2367 \\
Motus~\citep{bi2026motus}                          & -- & -- & w/o & 77.56 & 77.00 & 77.30  & 2367 \\
LingBot-VA~\citep{li2026lingbot-va}                & 25 & 50 & w/ & 92.90 & 91.50 & 92.20 & 7659 \\
LingBot-VA~\citep{li2026lingbot-va}                & 25 & 50 & w/o & 80.60 & - & 80.60 & 7659 \\
{Flash-WAM}~\citep{akbari2026flashwam}                           & 1 & 2 & w/o & {88.42}  & {82.66} & {85.54} & 707 \\
% \rowcolor{gray!15}
{Flash-WAM}~\citep{akbari2026flashwam}                                & 1 & 1 & w/o & {82.56}  & {80.26} & {81.41} & 609 \\
Efficient-WAM~\citep{li2026efficientwam} & 10 & 10 & w/o & 86.70 & 85.70 & 86.20 & 568 \\
Efficient-WAM-RT~\citep{li2026efficientwam} & 2 & 10 & w/o & 83.10 & 82.00 & 82.55 & 314 \\
% \rowcolor{gray!15}
\midrule
\multicolumn{8}{l}{\textit{Direct policy-based WAM methods}}\\
Fast-WAM~\citep{fastwam} & -- & 10 & w/o & 91.88 & 91.78 & 91.83 & 397 \\
Faster-WAM~\citep{ma2026fasterwam_huawei} & -- & 10 & w/o & 89.70 & 88.64 & 89.17 & -- \\
Light-WAM~\citep{li2026lightwam} & -- & -- & w/o & 76.40 & 76.30 & 76.40 & 107 \\
\midrule
\textbf{OSWAM} ({Ours}) & -- & 1 & w/o & 90.04 & 89.96 & 90.00 & 142 \\
\bottomrule
\end{tabular}%
}
% \vspace{-4mm}
\label{tab:main_results}
\end{table}

\subsection{Main Results}
\label{sec:main_results}
\textbf{NAVSIM}. Table~\ref{tab:navsim} compares OSWAM with state-of-the-art planners on NAVSIM.
OSWAM achieves a competitive PDMS of $90.3$ with a latency of $136$ ms.
Compared with SimWAM, OSWAM matches its PDMS while reducing action denoising from ten steps to one, yielding a $3.01\times$ speedup.
OSWAM also outperforms Metis by $1.2$ PDMS with a $3.24\times$ speedup.
Compared with DriveVA, OSWAM achieves a $4.36\times$ speedup with only a $0.2$ lower PDMS.
These results demonstrate the effectiveness of structured initialization in enabling efficient one-step action generation while maintaining competitive planning performance.

\textbf{RoboTwin 2.0}. Table~\ref{tab:main_results} reports success rates across 50 tasks under Clean and Randomized settings.
% Detailed per-task success rates under both settings are provided
% in Table~\ref{tab:robotwin-oswam}.
OSWAM achieves an average success rate of $90.00\%$ with a latency of $142$ms.
Compared with Fast-WAM, OSWAM reduces action denoising from ten steps to one and achieves a $2.80\times$ speedup, with only $1.83\%$ decrease in average success rate.
Compared with the one-step Flash-WAM variant, OSWAM improves average success by $8.59\%$ and achieves a $4.29\times$ speedup.
While Light-WAM achieves lower latency, OSWAM substantially improves the average success at the cost of only $35$ms additional latency.
Overall, OSWAM is competitive with multi-step direct-policy WAMs while substantially accelerating inference on embodied control tasks.

\textbf{Inference Efficiency}. Table~\ref{tab:inference_efficiency} compares inference efficiency with cached language context under the same environment.
On RoboTwin 2.0, OSWAM reduces action GPU time from $67.61$ to $6.78$ms compared with Fast-WAM, while imagination tokens increase visual GPU time by $5.48$ms.
This yields a $3.39\times$ overall speedup, reducing latency from $323.67$ to $95.35$ms with nearly unchanged peak memory.
On NAVSIM, OSWAM similarly achieves a $3.32\times$ speedup over SimWAM, reducing overall latency to $99.70$ms.
With compiled inference using \texttt{torch.compile} and CUDA Graph optimization, OSWAM achieves latencies of $40.81$ms on RoboTwin 2.0 and $52.64$ms on NAVSIM, supporting inference rates above $10$Hz and facilitating deployment in latency-critical applications such as autonomous driving.
These results show that the reduction in action denoising cost outweighs the additional computation introduced by imagination tokens.

\begin{table}[t]
  \caption{
  \textbf{Inference efficiency on RoboTwin 2.0 and Navsim. }
  Latency is measured with cached language context on a single NVIDIA RTX 5090 GPU and Intel Xeon Platinum 8470Q. All time are reported in milliseconds (ms).
  }
  \label{tab:inference_efficiency}
  \noindent
  \resizebox{\textwidth}{!}{%
  \begin{tabular}{llcccccc}
  \toprule
  Dataset & Model & VAE GPU Time & Visual GPU Time & Action GPU Time & Peak Mem. (GB) & Overall Latency  & Compiled Latency  \\
  \midrule
  % $\pi_{0.5}$~\cite{intelligence2025pi_}
  % & -- & -- & -- & $>8$GiB & -- & 76ms$^\ast$ \\
  \multirow{4}{*}{RoboTwin 2.0}
  & LingBot-VA
  & 32.65 & 1165.12 & 1205.38 & 19.12 & 3365.35 & -- \\
  & Motus
  & 11.14 & 567.37 & Shared & 17.95 & 1472.85 & -- \\
  & Fast-WAM
  & 5.27 & 18.45 & 67.61 & 12.70 & 323.67 & 73.64 \\
  % % \rowcolor[HTML]{F2F2F2} \textbf{Light-WAM}
  % & -- & -- & -- & -- & -- & -- \\
  & \textbf{OSWAM}
  & 5.27 & 23.93 & 6.78 & 12.75 & \textbf{95.35} & \textbf{40.81} \\
  \midrule
  \multirow{2}{*}{Navsim}
  & SimWAM
  & 9.83 & 22.80 & 72.78 & 12.87 & 330.8 & 90.53 \\
  & \textbf{OSWAM}
  & 9.82 & 30.87 & 7.32 & 12.90 & \textbf{99.70} & \textbf{52.64} \\
  \bottomrule
  \end{tabular}%
  }
  % \vspace{-6mm}
\end{table}

% %%%%% New ablation studies %%%%%
% \begin{table*}[t]
%   \caption{Ablations on the NAVSIM navtest.}
%   \centering
%   \begin{minipage}[t]{0.58\linewidth}
%      \centering
%      \setlength{\tabcolsep}{1mm}
%      \resizebox{0.78\columnwidth}{!}{
%      \begin{tabular}{lccccccc}
%      \toprule
%       &  \multicolumn{3}{c}{Navsim} &  \multicolumn{3}{c}{RoboTwin}\\
%     \multirow{-2}{*}{Method}& Latency & PDMS & ADE & Latency & Clean & Rand \\
%         \midrule
%         Baseline    &       & 71.04 &  1.049  &   & 57.38 & 55.80 \\
%         \quad + AIT &       & 89.81 & 0.594 &   & 87.42 & 88.16 \\
%         \quad + Distiallation&      & 89.96  &  0.589 &   & \\
%         \quad + SIT &       & 90.28 & 0.587 &    & 90.04 & 89.96\\
%         \bottomrule
%      \end{tabular}}
%      % \vspace*{2mm}
%   \end{minipage}\quad
%   \begin{minipage}[t]{0.32\linewidth}
%     \centering
%     \setlength{\tabcolsep}{1mm}
%     \resizebox{0.71\columnwidth}{!}{
%      \begin{tabular}{lcc}
%      \toprule
%      $\lambda_{\mathrm{dist}}$ & PDMS & ADE  \\
%      \midrule
%         Baseline                    & 89.81 & 0.594  \\
%         0.01  & 89.71 & 0.595\\
%         0.1   & 89.92 & 0.596  \\
%         1     & 90.01 & 0.591 \\
%         10  & \textbf{90.28} & \textbf{0.587}\\
%      \bottomrule
%      \end{tabular}
%      }
%      % \vspace*{2mm}
%   \end{minipage}
%   \label{tab:ablations}
%   \vspace*{-1mm}
% \end{table*}

%%%%% New ablation studies %%%%%
\begin{table*}[t]
    \centering
    \begin{minipage}[b]{0.66\linewidth}
        \centering
        \caption{Component ablations on NAVSIM and RoboTwin.}
        \label{tab:ablations}
        \setlength{\tabcolsep}{1mm}
        \renewcommand{\arraystretch}{1.1}
        \resizebox{\linewidth}{!}{%
            \begin{tabular}{lcccccc}
                \toprule
                \multirow{2}{*}{Method}
                & \multicolumn{3}{c}{NAVSIM}
                & \multicolumn{3}{c}{RoboTwin} \\
                \cmidrule(lr){2-4}\cmidrule(lr){5-7}
                & Latency$\downarrow$ & PDMS$\uparrow$ & ADE$\downarrow$
                & Latency$\downarrow$ & Clean$\uparrow$ & Rand.$\uparrow$ \\
                \midrule
                Baseline
                    & 108.35 & 71.04 & 1.049 & 112.84 & 57.38 & 55.80 \\
                \quad + AIT
                    & 135.92 & 89.81 & 0.594 & 139.82 & 87.42 & 88.16 \\
                \quad + Distillation
                    & 135.92 & 89.96 & 0.589 & 139.82 & 89.42 & 89.62 \\
                \quad + SIT
                    & 136.09 & \textbf{90.28} & \textbf{0.587}
                    & 142.67 & \textbf{90.04} & \textbf{89.96} \\
                \bottomrule
            \end{tabular}%
        }
    \end{minipage}%
    \hfill
    \begin{minipage}[b]{0.30\linewidth}
        \centering
        \caption{Ablation on $\lambda_{\mathrm{dist}}$.}
        \label{tab:dist_weight}
        \setlength{\tabcolsep}{1mm}
        \renewcommand{\arraystretch}{1.1}
        \resizebox{0.88\linewidth}{!}{%
            \begin{tabular}{lcc}
                \toprule
                $\lambda_{\mathrm{dist}}$
                    & PDMS$\uparrow$ & ADE$\downarrow$ \\
                \midrule
                Baseline & 89.81 & 0.594 \\
                0.01     & 89.71 & 0.595 \\
                0.1      & 89.92 & 0.596 \\
                1        & 90.01 & 0.591 \\
                10       & \textbf{90.28} & \textbf{0.587} \\
                \bottomrule
            \end{tabular}%
        }
    \end{minipage}
    % \vspace{-5mm}
\end{table*}

\subsection{Ablation Studies}

\textbf{Effectiveness of Each Component.}
To more clearly illustrate the effectiveness of the different components in OSWAM, we conduct experiments by adding each of them to a baseline, which is set to one-step Fast-WAM. As shown in Table~\ref{tab:ablations}, AIT substantially improves performance across both benchmarks, raising NAVSIM PDMS from $71.04$ to $89.81$, and increasing RoboTwin success rates from $57.38\%$ to $87.42\%$ under Clean settings and from $55.80\%$ to $88.16\%$ under Randomized settings. The gains demonstrate the importance of structured start points for one-step action generation. Besides, privileged-information distillation further improves PDMS to $89.96$ and RoboTwin success rates to $89.42\%$ and $89.62\%$, without increasing inference latency. Finally, incorporating SIT yields the best performance on both benchmarks, demonstrating the complementary benefit of aggregating scene context for learning effective action priors.

\textbf{Distillation Weight.}
Table~\ref{tab:dist_weight} studies the effect of the distillation weight $\lambda_{\mathrm{dist}}$ on NAVSIM.
While a small weight of $0.01$ slightly underperforms the baseline, increasing the distillation strength generally improves performance.
The best result is obtained with $\lambda_{\mathrm{dist}}=10$, reaching a $90.28$ PDMS and a $0.587$ ADE.
This suggests that sufficiently strong distillation supervision is important for transferring future context into AIT.
We therefore use $\lambda_{\mathrm{dist}}=10$ as the default setting.

%%%%% New ablation studies %%%%%
\begin{table*}[t]
    \centering
    \begin{minipage}[b]{0.48\linewidth}
        \centering
        \caption{Action initialization ablation.}
        \label{tab:renoising_timestep}
        \setlength{\tabcolsep}{2mm}
        \renewcommand{\arraystretch}{1.1}
        \resizebox{0.85\linewidth}{!}{%
            \begin{tabular}{lcc}
                \toprule
                $\tau_r$ & PDMS$\uparrow$ & ADE$\downarrow$ \\
                \midrule
                Rectified-flow refinement & 89.97 & 0.609 \\
                400            & 90.17 & 0.587 \\
                500            & 90.28 & 0.587 \\
                600            & 90.08 & 0.598 \\
                Uniform (400-600) & 90.02  &  0.591 \\
                \bottomrule
            \end{tabular}%
        }
    \end{minipage}%
    \hfill
    \begin{minipage}[b]{0.44\linewidth}
        \centering
        \caption{Teacher video timestep ablation.}
        \label{tab:teacher_video_timestep}
        \setlength{\tabcolsep}{2mm}
        \renewcommand{\arraystretch}{1.1}
        \resizebox{0.85\linewidth}{!}{%
            \begin{tabular}{lcc}
                \toprule
                Teacher video timestep
                    & PDMS$\uparrow$ & ADE$\downarrow$ \\
                \midrule
                0    & 90.20 & 0.585 \\
                200  & 90.28 & 0.587 \\
                400  & 90.19 & 0.588 \\
                600  & 90.16 & 0.590 \\
                800  & 90.16 & 0.591 \\
                % 1000 & 90.20 & 0.591 \\
                \bottomrule
            \end{tabular}%
        }
    \end{minipage}
    \vspace{-4mm}
\end{table*}

\textbf{Action Initialization.}
To evaluate different strategies for utilizing the structured action prior, we compare rectified-flow refinement~\citep{liu2022rectifiedflow} with re-noising at different noise levels.
As shown in Table~\ref{tab:renoising_timestep}, all re-noising variants outperform rectified-flow refinement, achieving higher PDMS and lower ADE.
Increasing $\tau_r$ from 400 to 500 improves PDMS without affecting ADE, whereas a further increase to 600 slightly degrades both metrics.
Uniformly sampling $\tau_r$ from 400 to 600 also underperforms the fixed setting of 500.
These results suggest that a fixed intermediate noise level provides a favorable trade-off between preserving the structured prior and allowing sufficient action refinement; we therefore set $\tau_r=500$ by default.

\textbf{Teacher Video Timestep.}
To investigate how the teacher noise level affects AIT distillation, we vary the teacher video timestep $\tau_v^T$ from 0 to 800.
As shown in Table~\ref{tab:teacher_video_timestep}, performance changes only slightly across the evaluated settings, with PDMS ranging from $90.16$ to $90.28$.
Even at $\tau_v^T=800$, where future video inputs are pure noise, OSWAM achieves a PDMS of $90.16$, close to the clean-future setting ($\tau_v^T=0$).
This suggests that the teacher's learned future-modeling representations can provide useful supervision for AIT even without access to clean ground-truth future frames.
We use $\tau_v^T=200$ by default, as it achieves the highest PDMS.
% Requires: booktabs, graphicx, multirow
%==========================================================================

\subsection{Real-World Evaluation}

% \begin{wraptable}{r}{0.46\textwidth}
%     \centering
%     \small
%     \caption{Real-world success rates (\%) on PiPER.}
%     \label{tab:real_world}

%     \setlength{\tabcolsep}{4pt}
%     \renewcommand{\arraystretch}{1.15}
%     \begin{tabular*}{\linewidth}
%         {@{\extracolsep{\fill}}lcc@{}}
%         \toprule
%         Task & Fast-WAM & OSWAM \\
%         \midrule
%         Bread into bowl & 80 & 85 \\
%         Eraser into drawer & 75 & 80 \\
%         \bottomrule
%     \end{tabular*}
% \end{wraptable}

\begin{wraptable}{r}{0.56\textwidth}
    \centering
    \small
    \setlength{\tabcolsep}{3pt}
    \renewcommand{\arraystretch}{1.12}
    \vspace{-6mm}
    \caption{Real-world evaluation on PiPER.
    SR: success rate (\%).
    Time: video-estimated completion time over
    successful episodes.}
    \label{tab:real_world}
    \begin{tabular*}{\linewidth}
        {@{\extracolsep{\fill}}lcccc@{}}
        \toprule
        & \multicolumn{2}{c}{Fast-WAM}
        & \multicolumn{2}{c}{OSWAM (Ours)} \\
        \cmidrule(lr){2-3}
        \cmidrule(l){4-5}
        Task
        & SR $\uparrow$ & Time (s) $\downarrow$
        & SR $\uparrow$ & Time (s) $\downarrow$ \\
        \midrule
        Bread into bowl
        & 80 & $24.6$
        & \textbf{85} & $\mathbf{16.2}$ \\
        Eraser into drawer
        & 75 & $17.6$
        & \textbf{80} & $\mathbf{13.8}$ \\
        \bottomrule
    \end{tabular*}
    \vspace{-4mm}
\end{wraptable}

We further evaluate OSWAM on a PiPER robotic arm using two real-world tabletop manipulation tasks: placing a packaged bread into a bowl and placing a board eraser into an open drawer. Both tasks require coordinated grasping, transport, and accurate placement, while the drawer task additionally imposes clearance constraints from the drawer opening and surrounding cabinet. We use 50 training demonstrations for bread and eraser placement, respectively.
Each method is evaluated over 20 trials per task. A trial is successful if the object is released into the target container and remains inside at the end of the execution.

\begin{figure*}[t!]
     \centering
    \includegraphics[width=0.9\textwidth]{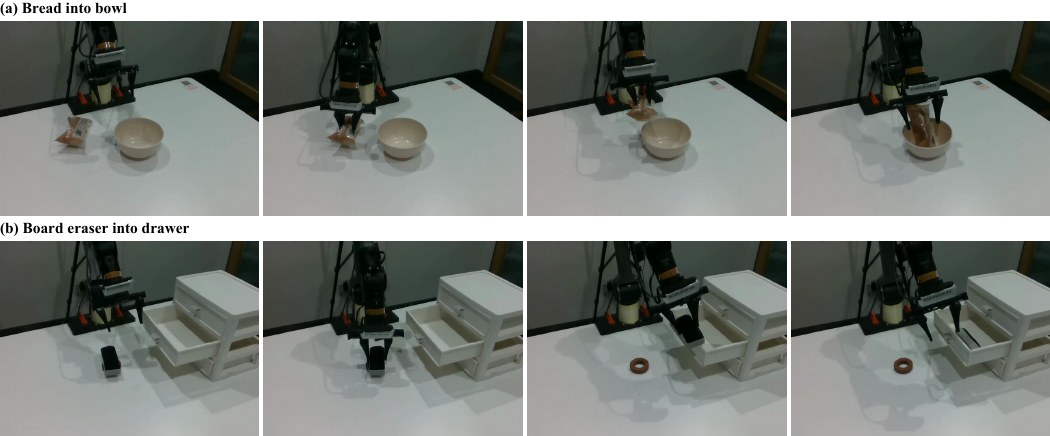}
     % \vspace{-5mm}
     \caption{Real-world manipulation with OSWAM on a PiPER robotic arm: (a) placing bread into a bowl and (b) placing a board eraser into an open drawer. 
     % Keyframes from left to right show the initial configuration, grasping, transport, and successful placement for each task.
     }
    \label{fig:real_world}
    % \vspace{-6mm}
\end{figure*}

As shown in Table~\ref{tab:real_world}, OSWAM achieves success rates of 85\% on bread placement and 80\% on eraser placement, compared with 80\% and 75\% for Fast-WAM, respectively. These results indicate that OSWAM's structured action initialization followed by a single Action DiT denoising step supports effective physical manipulation, with higher observed success rates than Fast-WAM on both evaluated tasks.
Among successful episodes, OSWAM reduces the mean video-estimated
task completion time by 34.2\% for bread placement and 21.8\% for eraser placement compared with Fast-WAM.
Figure~\ref{fig:real_world} presents representative successful rollouts, showing the progression from grasping and transport to final placement. OSWAM places the bread inside the bowl and releases the board eraser into the target drawer.
Together, the quantitative results and execution examples complement our simulation evaluation and demonstrate the applicability of OSWAM to real-world manipulation.
Additional execution videos are available at \url{https://gwenzhang.github.io/OSWAM}.

\section{Conclusion}
%==========================================================================
In this paper, we presented OSWAM, an efficient world-action model for one-step action generation.
To address the reliance of direct-policy WAMs on iterative action denoising, we introduced Action Imagination Tokens to learn a structured initialization, supported by Scene Imagination Tokens that aggregate complementary scene context.
We further developed privileged-information distillation to transfer full future context into the Action Imagination Tokens.
Together, these designs enable action generation with a single Video DiT pass and one Action DiT denoising step.
Experiments on NAVSIM and RoboTwin 2.0 demonstrated the competitive planning and manipulation performance of OSWAM and its substantial latency reductions over direct-policy baselines.
OSWAM offers a practical approach to efficient world-action modeling for latency-critical control.

\textbf{Limitations.} Our study is focused on learning action DiT structured initialization.
Although SIT and privileged-information distillation support this objective, their potential to directly enhance the Action DiT remains unexplored.
Extending scene-context conditioning and privileged supervision to the action expert is a promising direction for future work.

%==========================================================================
% Acknowledgements
%==========================================================================

%==========================================================================
% Bibliography
%==========================================================================
{\small
\bibliography{ref}
}

\appendix

\clearpage
\setcounter{table}{0}
\setcounter{equation}{0}
\setcounter{figure}{0}
\renewcommand{\thetable}{\thesection.\arabic{table}}
\renewcommand{\theequation}{\thesection.\arabic{equation}}
\renewcommand{\thefigure}{\thesection.\arabic{figure}}
\begin{center}
    {\LARGE\bfseries Appendix}
\end{center}

We provide the following materials in this appendix:
\begin{itemize}
    % \item Section~\ref{appedix:preliminary}: Preliminary for OSWAM.
    \item Section~\ref{appedix:nuscenes}: More experiment results.
    % \item Section~\ref{appedix:ablationstd}: More ablation studies.
    \item Section~\ref{appedix:visal}: Qualitative visualizations.
    % \item Section~\ref{appedix:llm}: Declaration of LLM assistance.
\end{itemize}

\section{More experiment results}
\label{appedix:nuscenes}

\subsection{More Implementation Details}
\textbf{nuScene.} We also evaluate OSWAM on the nuScenes~\citep{nuscenes} dataset for autonomous driving.
nuScenes comprises 1,000 real-world driving scenes, with keyframes annotated at 2 Hz.
The scene images are captured by 6 cameras covering 360 degree FOV horizontally.
The dataset is split into 700 scenes for training, 150 for validation, and 150 for testing.
Following previous works~\citep{zhao2026pwm,wang2025omnidrive}, $L_2$ Displacement Error (DE) and Collision Rate (CR) are adopted to comprehensively evaluate the planning performance.

For nuScenes, we use a single front camera at $384\times672$ resolution.
Each training sample contains the current frame and eight future frames sampled at 2Hz.
We use eight action steps to match the four-second video horizon, with supervision and evaluation limited to the first three seconds.
Each action is represented as $(x,y,\theta)$ in the current LiDAR coordinate frame, with padded and unsupervised positions excluded from the action loss.
For Role-aware 3D RoPE, we set $s_a=s_v=1$ and $d_t=4$, giving AIT and SIT temporal coordinates $t_n=n/4$ for $(n=1,\ldots,8)$.
Video latents use temporal coordinates $\{0,1,2\}$ during training.

We train for 10,000 steps with a global batch size of 128 using AdamW, a learning rate of $10^{-4}$, weight decay of $0.01$, and cosine decay with 5\% linear warmup.
For distillation, the frozen teacher uses a video timestep of 200, and feature alignment is applied after Video DiT blocks $[8,16,24,30]$.
We set $\lambda_{\mathrm{dist}}=10$.
Distillation starts after 500 steps, with its weight increased linearly over the next 500 steps.
We set the AIT diffusion timestep to $t_{\max}=1000$ and the re-noising timestep for Action DiT initialization to $\tau_r=500$ ($\sigma_r=0.5$).

\begin{table}[!b]
  \caption{Comparison with state-of-the-art planners on the NAVSIM-v2
  \texttt{navtest}. $N^{v}$ and $N^{a}$ denote the number of video
  and action denoising steps, respectively.}
  \label{tab:navsim_v2}
  \centering
  \setlength\tabcolsep{2.4mm}
  \renewcommand{\arraystretch}{1.0}
  \resizebox{0.9\linewidth}{!}{%
  \begin{tabular}{l c c c c c c c c c c c c}
    \toprule
    Method & $N^{v}$ & $N^{a}$
    & NC$\uparrow$ & DAC$\uparrow$ & DDC$\uparrow$
    & TLC$\uparrow$ & EP$\uparrow$ & TTC$\uparrow$
    & LK$\uparrow$ & HC$\uparrow$ & EC$\uparrow$
    & EPDMS$\uparrow$ \\
    \midrule
    Human Agent & -- & --
    & 100.0 & 100.0 & 99.8 & 100.0 & 87.4 & 100.0
    & 100.0 & 98.1 & 90.1 & 90.3 \\
    \midrule
    \multicolumn{13}{l}{\textit{Traditional E2E and VLA-based planners}} \\
    TransFuser~\citep{chitta2022transfuser} & -- & --
    & 96.9 & 89.9 & 97.8 & 99.7 & 87.1 & 95.4
    & 92.7 & 98.3 & 87.2 & 76.7 \\
    Hydra-MDP++~\citep{li2025hydramdp_pp} & -- & --
    & 97.2 & 97.5 & 99.4 & 99.6 & 83.1 & 96.5
    & 94.4 & 98.2 & 70.9 & 81.4 \\
    GTRS-Dense~\citep{li2025gtrs} & -- & --
    & 97.6 & 97.5 & 99.0 & 99.9 & 87.9 & 97.0
    & 95.9 & 97.5 & 55.9 & 82.3 \\
    DriveSuprim~\citep{yao2026drivesuprim} & -- & --
    & 97.5 & 96.5 & 99.4 & 99.6 & 88.4 & 96.6
    & 95.5 & 98.3 & 77.0 & 83.1 \\
    ARTEMIS~\citep{feng2025artemis} & -- & --
    & 98.3 & 95.1 & 98.6 & 99.8 & 81.5 & 97.4
    & 96.5 & 98.3 & -- & 83.1 \\
    DiffusionDrive~\citep{liao2025diffusiondrive} & -- & --
    & 98.2 & 95.9 & 99.4 & 99.8 & 87.5 & 97.3
    & 96.8 & 98.3 & 87.7 & 84.5 \\
    % World4Drive & -- & --
    % & 97.8 & 96.3 & 99.4 & 99.8 & 88.3 & 97.1
    % & 97.7 & 98.0 & 53.9 & 84.8 \\
    % Drive-JEPA & -- & --
    % & 98.8 & 97.4 & 99.0 & 99.8 & 83.5 & 98.0
    % & 96.2 & 98.1 & 85.6 & 85.4 \\
    % WorldRFT$^{*}$ & -- & --
    % & 97.8 & 96.5 & 99.5 & 99.8 & 88.5 & 97.0
    % & 97.4 & 98.1 & 69.1 & 86.7 \\
    ReCogDrive~\citep{xiong2026recogdrive} & -- & --
    & 98.3 & 95.2 & 99.5 & 99.8 & 87.1 & 97.5
    & 96.6 & 98.3 & 86.5 & 83.6 \\
    SGDrive~\citep{li2026sgdrive} & -- & --
    & 98.6 & 94.3 & 99.5 & 99.9 & 86.0 & 97.9
    & 96.1 & 98.3 & 85.9 & 86.2 \\
    % Vega & -- & --
    % & 98.9 & 95.3 & 99.4 & 99.9 & 87.0 & 98.4
    % & 96.1 & 98.3 & 76.3 & 86.9 \\
    % DriveFine$^{*}$ & -- & --
    % & 98.7 & 97.3 & 98.8 & 99.8 & 88.2 & 97.8
    % & 97.7 & 98.4 & 84.7 & 87.1 \\
    DriveVLA-W0~\citep{li2026drivevla-w0} & -- & --
    & 98.5 & 99.1 & 98.0 & 99.7 & 86.4 & 98.1
    & 93.2 & 97.9 & 58.9 & 86.1 \\
    \midrule
    \multicolumn{13}{l}{\textit{WAM-based planners}} \\
    Epona~\citep{zhang2025epona} & -- & 100
    & 97.1 & 95.7 & 99.3 & 99.7 & 88.6 & 96.3
    & 97.0 & 98.0 & 67.8 & 85.1 \\
    DriveLaW~\citep{xia2026drivelaw} & 1 & 5
    & 98.7 & 96.9 & 99.6 & 99.8 & 87.5 & 98.3
    & 97.6 & 98.4 & 77.4 & 88.6 \\
    Metis~\citep{li2026metis} & -- & 10
    & 98.4 & 97.2 & 99.6 & 99.8 & 87.8 & 97.7
    & 97.8 & 98.4 & 88.0 & 89.5 \\
    SimWAM~\citep{zhao2026simwam} & -- & 10
    & 98.6 & 98.0 & 99.7 & 99.9 & 87.5 & 98.4
    & 97.9 & 98.3 & 84.4 & 90.2 \\
    \midrule
    \textbf{OSWAM} (Ours) & -- & 1
    & 98.5 & 98.0 & 99.7 & 99.8 & 87.9 & 98.4
    & 98.0 & 98.4 & 84.3 & \textbf{90.3} \\
    \bottomrule
  \end{tabular}%
  }
  % \vspace{-3mm}
\end{table}

\begin{table*}[ht]
\setlength{\tabcolsep}{0.01\linewidth}
\caption{\textbf{End-to-end motion planning performance on the
nuScenes~\cite{nuscenes} dataset.}}
\centering
\resizebox{0.9\linewidth}{!}{%
\begin{tabular}{l|c|cccc|cccc}
\toprule
\multirow{2}{*}{Method}
& \multirow{2}{*}{Auxiliary Supervision}
& \multicolumn{4}{c|}{L2 (m) $\downarrow$}
& \multicolumn{4}{c}{Collision Rate (\%) $\downarrow$} \\
& & 1s & 2s & 3s & \cellcolor{gray!30}Avg.
  & 1s & 2s & 3s & \cellcolor{gray!30}Avg. \\
\midrule
ST-P3~\cite{hu2022stp3}
& Map \& Box \& Depth
& 1.33 & 2.11 & 2.90 & \cellcolor{gray!30}2.11
& 0.23 & 0.62 & 1.27 & \cellcolor{gray!30}0.71 \\
UniAD~\cite{hu2023uniad}
& {\footnotesize Map \& Box \& Motion \& Tracklets \& Occ}
& 0.48 & 0.96 & 1.65 & \cellcolor{gray!30}1.03
& 0.05 & 0.17 & 0.71 & \cellcolor{gray!30}0.31 \\
OccNet~\cite{tong2023occnet}
& 3D-Occ \& Map \& Box
& 1.29 & 2.13 & 2.99 & \cellcolor{gray!30}2.14
& 0.21 & 0.59 & 1.37 & \cellcolor{gray!30}0.72 \\
OccWorld~\cite{zheng2024occworld}
& 3D-Occ
& 0.52 & 1.27 & 2.41 & \cellcolor{gray!30}1.40
& 0.12 & 0.40 & 2.08 & \cellcolor{gray!30}0.87 \\
VAD-Tiny~\cite{jiang2023vad}
& Map \& Box \& Motion
& 0.60 & 1.23 & 2.06 & \cellcolor{gray!30}1.30
& 0.31 & 0.53 & 1.33 & \cellcolor{gray!30}0.72 \\
VAD-Base~\cite{jiang2023vad}
& Map \& Box \& Motion
& 0.54 & 1.15 & 1.98 & \cellcolor{gray!30}1.22
& 0.04 & 0.39 & 1.17 & \cellcolor{gray!30}0.53 \\
GenAD~\cite{zheng2024genad}
& Map \& Box \& Motion
& 0.36 & 0.83 & 1.55 & \cellcolor{gray!30}0.91
& 0.06 & 0.23 & 1.00 & \cellcolor{gray!30}0.43 \\
\midrule
Doe-1~\cite{zheng2024doe1}
& QA
& 0.50 & 1.18 & 2.11 & \cellcolor{gray!30}1.26
& 0.04 & 0.37 & 1.19 & \cellcolor{gray!30}0.53 \\
Epona~\citep{zhang2025epona}
& None
& 0.61 & 1.17 & 1.98 & \cellcolor{gray!30}1.25
& 0.01 & 0.22 & 0.85 & \cellcolor{gray!30}0.36 \\
DriveLaW~\citep{xia2026drivelaw}
& None
& 0.44 & 1.10 & 1.91 & \cellcolor{gray!30}1.15
& 0.15 & 0.10 & 0.48 & \cellcolor{gray!30}0.24 \\
PWM~\citep{zhao2026pwm}
& Text 
& 0.20 & 0.38 & \textbf{0.65} & \cellcolor{gray!30}\textbf{0.41}
& 0.01 & 0.02 & 0.09 & \cellcolor{gray!30}0.04 \\
\textbf{OSWAM} (Ours)
& None
& \textbf{0.19} & \textbf{0.36} & 0.66
& \cellcolor{gray!30}\textbf{0.41}
& \textbf{0.00} & \textbf{0.01} & \textbf{0.07}
& \cellcolor{gray!30}\textbf{0.03} \\
\bottomrule
\end{tabular}%
}
% \vspace{-6mm}
\label{tab:plan_nusc}
\end{table*}

\subsection{More Results}
\textbf{Navsim v2.} Compared with Navsim v1, Navsim v2 introduces the Extended Predictive Driver Model Score (EPDMS).
EPDMS incorporates Driving Direction Compliance (DDC), Traffic Light Compliance (TLC), Lane Keeping (LK), and Extended Comfort (EC), while extending the original comfort metric to History Comfort (HC).
As shown in Table~\ref{tab:navsim_v2}, OSWAM achieves the highest EPDMS of $90.3$ among the compared planners with only one action denoising step.
It slightly surpasses SimWAM $90.2$ and outperforms Metis $89.5$, both of which require ten action denoising steps.
These results further demonstrate the effectiveness of OSWAM under more comprehensive evaluation criteria encompassing safety, traffic-rule compliance, and driving comfort.

\textbf{nuScene}. Table~\ref{tab:plan_nusc} reports open-loop planning performance on nuScenes.
Without auxiliary supervision, OSWAM achieves an average $0.41$ m L2 error, matching PWM, while reducing the average collision rate from $0.04\%$ to $0.03\%$.
% Compared with PWM, OSWAM matches its average L2 error while reducing the average collision rate from $0.04\%$ to $0.03\%$.
It  also achieves the lowest L2 errors at 1 and 2 seconds and the lowest collision rates across all evaluated horizons.
These results demonstrate that OSWAM not only accelerates action generation but also delivers strong open-loop planning performance, matching the best average L2 error while achieving the lowest average collision rate.

\textbf{RoboTwin Per-task Success Rates}.
Table~\ref{tab:robotwin-oswam} reports per-task success rates across 50 manipulation tasks on RoboTwin 2.0.
% Compared with Fast-WAM, OSWAM reduces action denoising from ten steps to one and inference latency from $397$ ms to $142$ ms, achieving a $2.80\times$ speedup with an overall average success rate only $1.83$ percentage points lower.
OSWAM matches or exceeds Light-WAM on $49$ tasks under the clean setting and $46$ tasks under the randomized setting.
OSWAM also matches or exceeds Fast-WAM on $22$ tasks under the clean setting and $15$ tasks under the randomized setting.
Notably, OSWAM achieves substantial gains over Fast-WAM on Open Microwave under both settings, as well as on Move Stapler Pad, Hanging Mug, and Move Can Pot under the randomized setting.
% These results highlight OSWAM's ability to substantially accelerate inference while retaining competitive task-level performance relative to Fast-WAM.
Together with the $2.80\times$ inference speedup over Fast-WAM, these findings show that OSWAM maintains competitive task-level performance while substantially reducing inference latency.

\textbf{One-Step Action Generation.}
Tables~\ref{tab:one_step_navsim} and~\ref{tab:one_step_robotwin} compare OSWAM with direct policy baselines with a one-step denoising setting.
On NAVSIM, OSWAM achieves a PDMS of $90.3$, outperforming the reported one-step results of SimWAM, Metis, and DriveVA by $21.4$, $3.3$, and $77.1$ points, respectively.
On RoboTwin 2.0, OSWAM improves the average success rate from $56.59\%$ to $90.00\%$ over our one-step Fast-WAM baseline, with consistent gains under both Clean and Randomized settings.
These results demonstrate that OSWAM enables effective one-step action generation across both autonomous driving and embodied control.

% \section{Ablation Studies}
% \label{appedix:ablationstd}

% Requires: booktabs, graphicx
\begin{table*}[t]
    \centering
    \begin{minipage}[b]{0.4\linewidth}
        \centering
        \caption{One-step planning performance on NAVSIM.}
        \label{tab:one_step_navsim}
        \setlength{\tabcolsep}{2mm}
        \renewcommand{\arraystretch}{1.1}
        \resizebox{0.95\linewidth}{!}{%
            \begin{tabular}{lccc}
                \toprule
                Method & PDMS$\uparrow$ & NC$\uparrow$ & DAC$\uparrow$ \\
                \midrule
                SimWAM (1-step)  & 68.9 & 97.4 & 91.3 \\
                Metis (1-step)   & 87.0 & --   & --   \\
                DriveVA (1-step) & 13.2 & 61.8 & 36.9 \\
                \midrule
                \textbf{OSWAM (Ours)}
                & \textbf{90.3} & \textbf{98.6} & \textbf{98.0} \\
                \bottomrule
            \end{tabular}%
        }
    \end{minipage}%
    \hfill
    \begin{minipage}[b]{0.54\linewidth}
        \centering
        \caption{One-step success rates on RoboTwin 2.0.}
        \label{tab:one_step_robotwin}
        \setlength{\tabcolsep}{2mm}
        \renewcommand{\arraystretch}{1.1}
        \resizebox{0.95\linewidth}{!}{%
            \begin{tabular}{lccc}
                \toprule
                Method & Clean$\uparrow$ & Rand.$\uparrow$ & Average$\uparrow$ \\
                \midrule
                Fast-WAM (1-step)
                & 57.38 & 55.80 & 56.59 \\
                \midrule
                \textbf{OSWAM (Ours)}
                & \textbf{90.04} & \textbf{89.96} & \textbf{90.00} \\
                \bottomrule
            \end{tabular}%
        }
    \end{minipage}
    \vspace{-5mm}
\end{table*}

\vspace{-6mm}
\section{Qualitative visualizations}
\label{appedix:visal}

\begin{figure*}[t!]
     \centering
     % \label{fig:comparison_appendix}
    \includegraphics[width=0.85\textwidth]{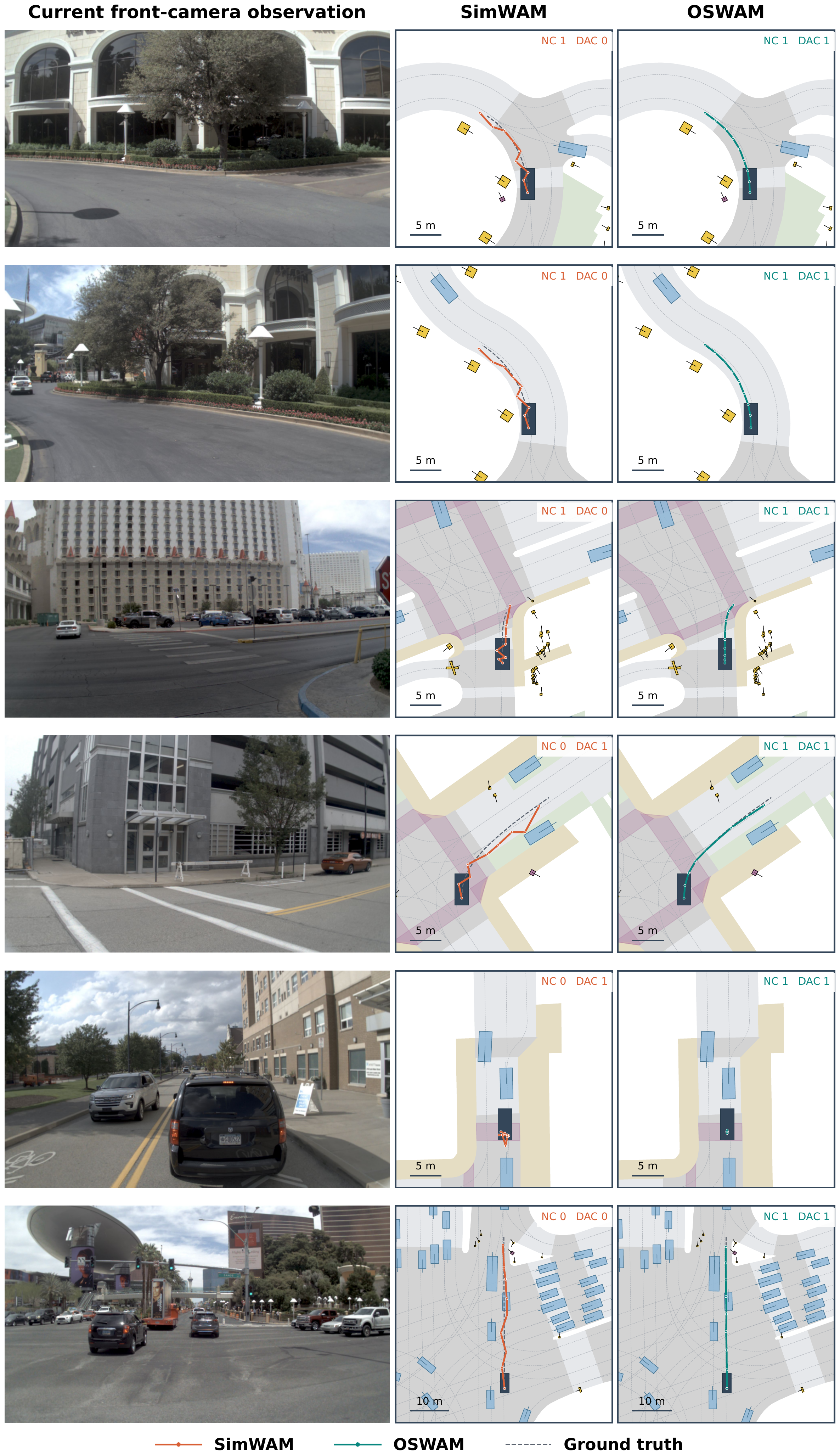}
    % \vspace{-4mm}
     \caption{Qualitative comparison of \textbf{one-step} SimWAM and OSWAM on Navsim v1.
     }
     \label{fig:comparison_appendix}
\end{figure*}

Figure~\ref{fig:comparison_appendix} compares SimWAM and OSWAM under the one-step action generation setting.
SimWAM struggles to produce smooth trajectories, particularly when turning or navigating complex interactions with surrounding agents, leading to drivable-area violations and at-fault collisions.
In contrast, OSWAM handles these challenging examples with a single action denoising step, producing smoother trajectories while satisfying both no-at-fault collision and drivable-area compliance criteria.
Together with the real-time inference speed reported in Table~\ref{tab:inference_efficiency}, these results highlight the potential of OSWAM for practical deployment in autonomous driving.

\begin{table*}[p]
\centering

% 将标题和表格作为整体，必要时等比例缩小至一页以内
\begin{adjustbox}{
    max totalsize={\textwidth}{0.98\textheight},
    center
}
\begin{minipage}{\textwidth}
\centering

\caption{Per-task success rates (\%) on RoboTwin 2.0 under clean and randomized settings.}
\label{tab:robotwin-oswam}

\footnotesize
\setlength{\tabcolsep}{2pt}
\renewcommand{\arraystretch}{1.05}

\begin{tabularx}{\linewidth}{
    @{}>{\raggedright\arraybackslash}X*{12}{r}@{}
}
\toprule
\multirow{2}{*}{Task} &
\multicolumn{2}{c}{\textbf{OSWAM}} &
\multicolumn{2}{c}{Fast-WAM} &
\multicolumn{2}{c}{Light-WAM} &
\multicolumn{2}{c}{LingBot-VA} &
\multicolumn{2}{c}{Motus} &
\multicolumn{2}{c}{$\pi_{0.5}$} \\
\cmidrule(lr){2-3}\cmidrule(lr){4-5}\cmidrule(lr){6-7}
\cmidrule(lr){8-9}\cmidrule(lr){10-11}\cmidrule(lr){12-13}
& Clean & Rand. & Clean & Rand. & Clean & Rand.
& Clean & Rand. & Clean & Rand. & Clean & Rand. \\
\midrule

Adjust Bottle & 100 & 99 & 100 & 100 & 100 & 100 & 90 & 94 & 89 & 93 & 100 & 99 \\
Beat Block Hammer & 99 & 92 & 99 & 97 & 83 & 80 & 96 & 98 & 95 & 88 & 96 & 93 \\
Blocks Ranking RGB & 99 & 97 & 100 & 100 & 96 & 91 & 99 & 98 & 99 & 97 & 92 & 85 \\
Blocks Ranking Size & 84 & 81 & 94 & 98 & 57 & 54 & 94 & 96 & 75 & 63 & 49 & 26 \\
Click Alarmclock & 100 & 99 & 100 & 100 & 100 & 100 & 99 & 100 & 100 & 100 & 98 & 89 \\
Click Bell & 97 & 99 & 100 & 100 & 100 & 100 & 100 & 100 & 100 & 100 & 99 & 66 \\
Dump Bin Bigbin & 92 & 95 & 97 & 96 & 81 & 75 & 89 & 96 & 95 & 91 & 92 & 97 \\
Grab Roller & 100 & 100 & 100 & 100 & 100 & 98 & 100 & 100 & 100 & 100 & 100 & 100 \\
Handover Block & 84 & 79 & 95 & 81 & 71 & 59 & 99 & 78 & 86 & 73 & 66 & 57 \\
Handover Mic & 96 & 100 & 99 & 100 & 90 & 94 & 94 & 96 & 78 & 63 & 98 & 97 \\
Hanging Mug & 47 & 68 & 58 & 62 & 25 & 17 & 40 & 28 & 38 & 38 & 18 & 17 \\
Lift Pot & 100 & 98 & 100 & 100 & 93 & 93 & 100 & 99 & 96 & 99 & 96 & 85 \\
Move Can Pot & 74 & 94 & 90 & 88 & 57 & 74 & 94 & 97 & 34 & 74 & 51 & 55 \\
Move Pillbottle Pad & 97 & 96 & 100 & 99 & 69 & 74 & 99 & 99 & 93 & 96 & 84 & 61 \\
Move Playingcard Away & 99 & 99 & 100 & 100 & 93 & 92 & 100 & 99 & 100 & 96 & 96 & 84 \\
Move Stapler Pad & 69 & 72 & 77 & 64 & 26 & 34 & 91 & 79 & 83 & 85 & 56 & 42 \\
Open Laptop & 99 & 98 & 98 & 100 & 91 & 97 & 92 & 94 & 95 & 91 & 90 & 96 \\
Open Microwave & 82 & 76 & 62 & 45 & 76 & 59 & 82 & 86 & 95 & 91 & 34 & 77 \\
Pick Diverse Bottles & 80 & 80 & 80 & 85 & 61 & 57 & 89 & 82 & 90 & 91 & 81 & 71 \\
Pick Dual Bottles & 97 & 92 & 100 & 96 & 90 & 63 & 100 & 99 & 96 & 90 & 93 & 63 \\
Place A2B Left & 89 & 95 & 95 & 93 & 84 & 83 & 97 & 93 & 88 & 79 & 87 & 82 \\
Place A2B Right & 96 & 94 & 93 & 99 & 89 & 85 & 97 & 95 & 91 & 87 & 87 & 84 \\
Place Bread Basket & 96 & 96 & 91 & 93 & 81 & 80 & 97 & 95 & 91 & 94 & 77 & 64 \\
Place Bread Skillet & 92 & 85 & 90 & 93 & 92 & 82 & 95 & 90 & 86 & 83 & 85 & 66 \\
Place Burger Fries & 97 & 99 & 96 & 99 & 95 & 98 & 97 & 95 & 98 & 98 & 94 & 87 \\
Place Can Basket & 79 & 65 & 71 & 69 & 57 & 57 & 81 & 84 & 81 & 76 & 62 & 62 \\
Place Cans Plasticbox & 97 & 94 & 99 & 96 & 37 & 68 & 100 & 99 & 98 & 94 & 94 & 84 \\
Place Container Plate & 99 & 99 & 96 & 100 & 99 & 94 & 99 & 97 & 98 & 99 & 99 & 95 \\
Place Dual Shoes & 83 & 87 & 94 & 88 & 53 & 51 & 94 & 89 & 93 & 87 & 75 & 75 \\
Place Empty Cup & 98 & 100 & 100 & 100 & 91 & 95 & 100 & 100 & 99 & 98 & 100 & 99 \\
Place Fan & 95 & 89 & 96 & 96 & 77 & 74 & 99 & 93 & 91 & 87 & 87 & 85 \\
Place Mouse Pad & 83 & 83 & 83 & 89 & 58 & 62 & 93 & 96 & 66 & 68 & 60 & 39 \\
Place Object Basket & 86 & 78 & 89 & 88 & 81 & 72 & 91 & 88 & 81 & 87 & 80 & 76 \\
Place Object Scale & 88 & 90 & 90 & 97 & 72 & 74 & 96 & 95 & 88 & 85 & 86 & 80 \\
Place Object Stand & 91 & 94 & 90 & 94 & 78 & 85 & 99 & 96 & 98 & 97 & 91 & 85 \\
Place Phone Stand & 99 & 97 & 97 & 99 & 85 & 88 & 97 & 97 & 87 & 86 & 81 & 81 \\
Place Shoe & 93 & 96 & 96 & 99 & 84 & 87 & 98 & 98 & 99 & 97 & 92 & 93 \\
Press Stapler & 80 & 86 & 90 & 97 & 65 & 76 & 85 & 82 & 93 & 98 & 87 & 83 \\
Put Bottles Dustbin & 90 & 89 & 95 & 90 & 65 & 65 & 87 & 91 & 81 & 79 & 84 & 79 \\
Put Object Cabinet & 84 & 85 & 94 & 89 & 80 & 68 & 85 & 87 & 88 & 71 & 80 & 79 \\
Rotate QRcode & 83 & 84 & 93 & 89 & 72 & 84 & 96 & 91 & 89 & 73 & 89 & 87 \\
Scan Object & 85 & 81 & 89 & 92 & 60 & 52 & 96 & 91 & 67 & 66 & 72 & 65 \\
Shake Bottle Horizontally & 100 & 100 & 100 & 100 & 100 & 98 & 100 & 99 & 100 & 98 & 99 & 99 \\
Shake Bottle & 100 & 100 & 100 & 100 & 100 & 99 & 100 & 97 & 100 & 97 & 99 & 97 \\
Stack Blocks Three & 94 & 95 & 95 & 97 & 65 & 67 & 99 & 98 & 91 & 95 & 91 & 76 \\
Stack Blocks Two & 99 & 100 & 100 & 100 & 94 & 91 & 100 & 98 & 100 & 98 & 97 & 100 \\
Stack Bowls Three & 84 & 75 & 80 & 81 & 65 & 72 & 86 & 83 & 79 & 87 & 77 & 71 \\
Stack Bowls Two & 95 & 96 & 92 & 98 & 91 & 97 & 94 & 98 & 98 & 98 & 95 & 96 \\
Stamp Seal & 92 & 87 & 90 & 94 & 60 & 63 & 96 & 97 & 93 & 92 & 79 & 55 \\
Turn Switch & 60 & 65 & 61 & 59 & 33 & 39 & 44 & 45 & 84 & 78 & 62 & 54 \\

\midrule
\textbf{Average} & 90.0 & 90.0 & 91.9 & 91.8 & 76.4 & 76.3
& 92.9 & 91.5 & 88.7 & 87.0 & 82.7 & 76.8 \\
\bottomrule
\end{tabularx}

\end{minipage}
\end{adjustbox}
\end{table*}

\end{document}